\documentclass[11pt]{article}

\usepackage[preprint]{acl}

\usepackage{times}
\usepackage{latexsym}
\usepackage{amssymb}
\usepackage[T1]{fontenc}

\usepackage[utf8]{inputenc}
\usepackage{amsmath}
\usepackage{booktabs}
\usepackage{microtype}

\usepackage{inconsolata}

\usepackage{graphicx}
   \usepackage{tikz}
   \usetikzlibrary{arrows.meta,positioning,fit,calc}
  
\usepackage{comment}

\usepackage{xcolor}
\usepackage{fvextra}                      
\usepackage[breakable,skins]{tcolorbox}   

\definecolor{promptframe}{HTML}{3B5B7A}
\definecolor{promptback}{HTML}{F6F8FA}
\definecolor{promptph}{HTML}{B03A2E}
\usepackage{xspace} 
\newcommand{\anonymousdataset}{Non-Public\xspace}
\newcommand{\panta}{Panta\xspace}

\newcommand{\ph}[1]{{\color{promptph}\textlangle\textit{#1}\textrangle}}

\newtcolorbox[auto counter]{promptbox}[3][]{%
  enhanced, breakable, colback=promptback, colframe=promptframe,
  boxrule=0.5pt, arc=2pt, left=6pt, right=6pt, top=4pt, bottom=4pt,
  fonttitle=\bfseries\small, coltitle=white,
  title={Prompt~\thetcbcounter: #2}, label={#3}, #1}

\fvset{fontsize=\footnotesize, breaklines=true, breakautoindent=true,
       breakindent=0pt, breaksymbolleft={}, commandchars=\\\|\^}

\title{FRAAC: A French Multimodal Large Language Model and Dataset for Pictogram-Based Augmentative and Alternative Communication}

\author{
  \textbf{Alexandre Audibert\textsuperscript{1}},
  \textbf{Maja Jonck Hjuler\textsuperscript{1}},
  \textbf{Benoit Favre\textsuperscript{2}},
  \textbf{François Portet\textsuperscript{1}}
\\
  \textsuperscript{1}Univ. Grenoble Alpes, CNRS, Grenoble INP, LIG, 38000 Grenoble, France
\\
  \textsuperscript{2}NAVER LABS Europe, Meylan, France
\\
  \small{
    \textbf{Correspondence:} \href{mailto:alexandre.audibert@univ-grenoble-alpes.fr}{alexandre.audibert@univ-grenoble-alpes.fr}
  }
}

\begin{document}
\maketitle
\begin{abstract}
Pictograms are a core modality in augmentative and alternative communication (AAC), yet recent multimodal large language models (MLLMs) cannot reliably interpret or produce pictogram sequences.
Two obstacles explain this gap: the absence of a standardized pictogram grammar and the scarcity of parallel text-pictogram data.
We introduce Retrieval-Augmented Pictogram Selection (RAPS), an annotation method that refines a rule-based transcription system's output with LLM-based concept extraction and retrieval over a pictogram vocabulary, handling cases where hand-written rules fail (named entities, numbers, polysemous words) without requiring new rules.  
Applied to French, RAPS yields a synthetic corpus of 1.4 million parallel pairs, of which 300k are publicly released. 
In a human ranking study where annotators order four pictogram transcriptions of the same sentence, RAPS outperforms the rule-based state of the art.
Moreover, we show that standard reference-based metrics separate systems of clearly different quality but fail to rank strong systems against each other.
Finally, we build an instruction-tuned model that supports interaction in several AAC modalities: it accepts pictogram or text input and answers in pictograms, speech, or text.
\end{abstract}

\section{Introduction}
\label{sec:intro}
Augmentative and Alternative Communication (AAC) refers to tools and strategies that help people with complex communication needs, such as those related to autism, intellectual disability, or acquired conditions like post-stroke aphasia \cite{Beukelman_book_2020}. 
AAC widely relies on pictograms, visual representations of concepts that support both language production and comprehension.
While multimodal large language models (MLLMs) can process or generate text, images, and speech \cite{xu2025qwen25omni}, they still do not support pictogram-based communication.
Doing so would enable automatic translation between pictograms and conventional language, facilitating interactions between AAC users and non-AAC users and helping bridge the digital divide.


\definecolor{pictoFill}{HTML}{FBEEE8}
\definecolor{pictoLine}{HTML}{C2562F}
\definecolor{pictoText}{HTML}{6E2A12}
\definecolor{speechFill}{HTML}{EAF0F7}
\definecolor{speechLine}{HTML}{3E6EA3}
\definecolor{speechText}{HTML}{22415F}
\definecolor{coreFill}{HTML}{ECEAFA}
\definecolor{coreLine}{HTML}{6F67C9}
\definecolor{coreText}{HTML}{342D78}
\definecolor{neutFill}{HTML}{F2F1EC}
\definecolor{neutLine}{HTML}{8A8983}
\definecolor{neutText}{HTML}{2B2B29}
\definecolor{errLine}{HTML}{C0392B}
\definecolor{okLine}{HTML}{2E8B57}

\begin{figure}[t]
  \centering
  \resizebox{0.9\columnwidth}{!}{%
  \begin{tikzpicture}[
    x=1mm, y=1mm,
    font=\footnotesize,
    every node/.style={inner sep=0pt, outer sep=0pt},
    panel/.style={draw=neutLine!70, line width=0.4pt, rounded corners=2.5pt, fill=white},
    ptitle/.style={anchor=north west, font=\scriptsize\bfseries, text=neutText},
    rowlab/.style={anchor=east, font=\scriptsize\bfseries, text=neutText},
    chip/.style={draw=pictoLine, line width=0.4pt, fill=white, rounded corners=1pt,
                 minimum width=6.4mm, minimum height=6.4mm},
    misschip/.style={chip, draw=errLine, dashed, line width=0.6pt},
    pcap/.style={font=\tiny, text=pictoText},
    ico/.style={draw=pictoLine, line width=0.4pt},
    core/.style={draw=coreLine, line width=0.7pt, fill=coreFill, rounded corners=1.5pt,
                 text=coreText, align=center, inner sep=1.5pt},
    neut/.style={draw=neutLine, line width=0.4pt, fill=neutFill, rounded corners=1.5pt,
                 text=neutText, align=center, inner sep=1.2pt},
    corpus/.style={draw=pictoLine, line width=0.6pt, fill=pictoFill, rounded corners=1.5pt,
                   text=pictoText, align=center, inner sep=1.5pt},
    flow/.style={-{Stealth[length=1.5mm, width=1.2mm]}, draw=neutLine, line width=0.45pt},
    inner/.style={-{Stealth[length=1.3mm, width=1.1mm]}, draw=coreLine!80, line width=0.45pt},
    side/.style={font=\tiny\itshape, text=neutText!80},
  ]

  \newcommand{\icoGrandma}[1]{%
    \draw[ico, fill=pictoFill] ([shift={(-0.7,-0.1)}]#1) -- ([shift={(0.7,-0.1)}]#1)
       -- ([shift={(1.4,-2.3)}]#1) -- ([shift={(-1.4,-2.3)}]#1) -- cycle;
    \draw[ico, fill=white] ([shift={(0,0.9)}]#1) circle (0.95mm);
    \draw[ico, fill=pictoLine!35] ([shift={(0,2.15)}]#1) circle (0.45mm);}
  \newcommand{\icoBake}[1]{%
    \draw[ico, fill=pictoFill] ([shift={(-1.7,1.0)}]#1) -- ([shift={(-1.7,-0.5)}]#1)
       .. controls ([shift={(-1.7,-1.1)}]#1) .. ([shift={(-1.1,-1.1)}]#1)
       -- ([shift={(1.1,-1.1)}]#1)
       .. controls ([shift={(1.7,-1.1)}]#1) .. ([shift={(1.7,-0.5)}]#1)
       -- ([shift={(1.7,1.0)}]#1) -- cycle;
    \draw[ico] ([shift={(-2.3,0.6)}]#1) -- ([shift={(-1.7,0.6)}]#1);
    \draw[ico] ([shift={(1.7,0.6)}]#1) -- ([shift={(2.3,0.6)}]#1);
    \draw[ico] ([shift={(-1.9,1.3)}]#1) -- ([shift={(1.9,1.3)}]#1);
    \draw[ico] ([shift={(0,1.3)}]#1) -- ([shift={(0,1.8)}]#1);
    \foreach \dx in {-1.0,0,1.0}
      \draw[ico, fill=pictoLine!45, line width=0.3pt] ([shift={(\dx,-2.5)}]#1)
         .. controls ([shift={(\dx-0.45,-2.0)}]#1) .. ([shift={(\dx,-1.5)}]#1)
         .. controls ([shift={(\dx+0.45,-2.0)}]#1) .. ([shift={(\dx,-2.5)}]#1);}
  \newcommand{\icoWith}[1]{%
    \draw[ico, fill=pictoFill] ([shift={(-2.3,-0.75)}]#1) rectangle ([shift={(-0.8,0.75)}]#1);
    \draw[ico, fill=pictoFill] ([shift={(0.8,-0.75)}]#1) rectangle ([shift={(2.3,0.75)}]#1);
    \draw[ico, line width=0.6pt] ([shift={(-0.8,0)}]#1) -- ([shift={(0.8,0)}]#1);}
  \newcommand{\icoOne}[1]{%
    \node[font=\large\bfseries, text=pictoText] at (#1) {1};}
  \newcommand{\icoPie}[1]{%
    \draw[ico, fill=pictoFill] ([shift={(-2.2,-1.0)}]#1) -- ([shift={(2.2,-1.0)}]#1)
       -- ([shift={(1.9,-0.3)}]#1)
       .. controls ([shift={(1.3,1.5)}]#1) and ([shift={(-1.3,1.5)}]#1) .. ([shift={(-1.9,-0.3)}]#1)
       -- cycle;
    \draw[ico, line width=0.3pt] ([shift={(-1.2,0.1)}]#1) -- ([shift={(-0.4,0.75)}]#1);
    \draw[ico, line width=0.3pt] ([shift={(-0.3,0.1)}]#1) -- ([shift={(0.5,0.75)}]#1);
    \draw[ico, line width=0.3pt] ([shift={(0.6,0.1)}]#1) -- ([shift={(1.2,0.6)}]#1);
    \draw[ico] ([shift={(-2.2,-1.0)}]#1) -- ([shift={(-1.9,-1.6)}]#1) -- ([shift={(1.9,-1.6)}]#1) -- ([shift={(2.2,-1.0)}]#1);}
  \newcommand{\icoApple}[1]{%
    \draw[ico, fill=pictoFill] ([shift={(0,0.8)}]#1)
       .. controls ([shift={(-0.8,1.5)}]#1) and ([shift={(-2.1,0.9)}]#1) .. ([shift={(-1.8,-0.6)}]#1)
       .. controls ([shift={(-1.5,-2.0)}]#1) and ([shift={(-0.5,-2.2)}]#1) .. ([shift={(0,-1.8)}]#1)
       .. controls ([shift={(0.5,-2.2)}]#1) and ([shift={(1.5,-2.0)}]#1) .. ([shift={(1.8,-0.6)}]#1)
       .. controls ([shift={(2.1,0.9)}]#1) and ([shift={(0.8,1.5)}]#1) .. ([shift={(0,0.8)}]#1);
    \draw[ico] ([shift={(0,0.8)}]#1) -- ([shift={(0.25,1.9)}]#1);
    \draw[ico, fill=okLine!35, draw=okLine] ([shift={(0.3,1.5)}]#1)
       .. controls ([shift={(0.9,2.2)}]#1) .. ([shift={(1.5,1.7)}]#1)
       .. controls ([shift={(0.9,1.2)}]#1) .. ([shift={(0.3,1.5)}]#1);}
  \newcommand{\icoClock}[3]{
    \draw[ico, fill=white] (#1) circle (1.85mm);
    \foreach \a in {0,90,180,270}
      \draw[ico, line width=0.3pt] ([shift={(\a:1.45mm)}]#1) -- ([shift={(\a:1.75mm)}]#1);
    \draw[ico, line width=0.5pt] (#1) -- ([shift={(#2:1.3mm)}]#1);
    \draw[ico, line width=0.6pt] (#1) -- ([shift={(#3:0.85mm)}]#1);
    \fill[pictoLine] (#1) circle (0.25mm);}
  \newcommand{\icoEat}[1]{%
    \draw[ico] ([shift={(-1.0,-2.2)}]#1) -- ([shift={(-1.0,0.4)}]#1);
    \draw[ico] ([shift={(-1.7,0.4)}]#1) -- ([shift={(-0.3,0.4)}]#1);
    \foreach \dx in {-1.7,-1.0,-0.3}
      \draw[ico] ([shift={(\dx,0.4)}]#1) -- ([shift={(\dx,2.0)}]#1);
    \draw[ico, fill=pictoFill] ([shift={(0.9,-2.2)}]#1) -- ([shift={(0.9,2.0)}]#1)
       .. controls ([shift={(1.9,1.4)}]#1) .. ([shift={(1.7,-0.1)}]#1) -- ([shift={(0.9,-0.1)}]#1);}

  \draw[panel] (0,69.5) rectangle (80,100);
  \node[ptitle] at (2,98.6) {(a) Text-to-pictogram transcription};
  \node[anchor=west, font=\scriptsize\itshape, text=neutText] at (4,94.4)
    {``Grandma bakes a pie with apples.''};

  \node[rowlab] at (20.5,88.5) {Rule-based};
  \foreach \x/\n in {25.5/r1, 33/r2, 40.5/r3, 48/r4, 55.5/r5} \node[chip] at (\x,88.5) (\n) {};
  \node[misschip] at (63,88.5) (r6) {};
  \icoGrandma{r1.center}
  \icoBake{r2.center}
  \icoOne{r3.center}
  \icoPie{r4.center}
  \icoWith{r5.center}
  \node[font=\small\bfseries, text=errLine] at (r6.center) {?};
  \draw[errLine, line width=0.8pt] (68.6,87.1) -- (71.4,89.9);
  \draw[errLine, line width=0.8pt] (68.6,89.9) -- (71.4,87.1);
  \foreach \x/\t in {25.5/grandma, 33/bake, 40.5/one, 48/pie, 55.5/with}
    \node[pcap] at (\x,83.6) {\t};
  \node[pcap, text=errLine] at (63,83.6) {apple};

  \node[rowlab] at (20.5,77.2) {RAPS};
  \foreach \x/\n in {25.5/s1, 33/s2, 40.5/s3, 48/s4, 55.5/s5} \node[chip] at (\x,77.2) (\n) {};
  \icoGrandma{s1.center}
  \icoBake{s2.center}
  \icoPie{s3.center}
  \icoWith{s4.center}
  \icoApple{s5.center}
  \fill[okLine] (s5.north east) circle (0.9mm);
  \draw[white, line width=0.5pt] ([shift={(-0.5,0)}]s5.north east) -- ([shift={(0.5,0)}]s5.north east);
  \draw[white, line width=0.5pt] ([shift={(0,-0.5)}]s5.north east) -- ([shift={(0,0.5)}]s5.north east);
  \draw[okLine, line width=0.9pt] (61.5,77.3) -- (62.8,75.9) -- (65.3,79.0);
  \foreach \x/\t in {25.5/grandma, 33/bake, 40.5/pie, 48/with, 55.5/apple}
    \node[pcap] at (\x,72.3) {\t};

  \draw[flow] (40,69.5) -- (40,63.5);
  \node[side, anchor=west] at (41.2,66.5) {applied at scale};

  \draw[panel] (0,39.5) rectangle (80,63.5);
  \node[ptitle] at (2,62.1) {(b) Pre-training corpus};

  \draw[draw=neutLine, line width=0.4pt, fill=neutFill, rounded corners=1.5pt]
    (3.5,40.9) rectangle (34.5,58.3);
  \node[anchor=west, font=\tiny\itshape, text=neutText!80] at (5,56.6) {Source};
  \node[anchor=east, font=\tiny\itshape, text=neutText!80] at (33,56.6) {\# pairs};
  \draw[neutLine!60, line width=0.3pt] (5,55.3) -- (33,55.3);
  \foreach \y/\name/\count in {
      53.8/Vikidia/90k,
      51.0/Cosmopedia/216k,
      48.2/\anonymousdataset/1.04M,
      45.4/Project Gutenberg/13k,
      42.6/Single pictograms/40k} {
    \node[anchor=west, font=\tiny, text=neutText] at (5,\y) {\name};
    \node[anchor=east, font=\tiny, text=neutText] at (33,\y) {\count};
  }

  \node[core, minimum width=12mm, minimum height=7mm] at (44,54.2) (raps) {\textbf{RAPS}};
  \node[corpus, minimum width=19mm, minimum height=8mm] at (67,54.2) (corp)
    {\textbf{1.4M pairs}\\[1pt]{\tiny (300k released)}};
  \node[core, minimum width=19mm, minimum height=6.5mm] at (67,43.6) (panta1) {\textbf{\panta-v1}};
  \draw[flow] (34.5,54.2) -- (raps.west);
  \draw[flow] (raps.east) -- (corp.west);
  \draw[flow] (corp.south) -- (panta1.north);
  \node[side, anchor=west] at (68.2,48.5) {trains};

  \draw[flow] (40,39.5) -- (40,33.5);
  \node[side, anchor=west] at (41.2,36.5) {initializes};

  \draw[panel] (0,1.5) rectangle (80,33.5);
  \node[ptitle] at (2,32.1) {(c) Question answering for AAC users};

  \node[rowlab] at (17,24.5) {Question};
  \node[chip] at (23,24.5) (q1) {};
  \node[chip] at (31,24.5) (q2) {};
  \icoClock{q1.center}{90}{40}
  \node[font=\scriptsize\bfseries, text=speechLine] at ([shift={(0.3,0.6)}]q1.north east) {?};
  \icoEat{q2.center}
  \node[pcap] at (23,19.6) {when};
  \node[pcap] at (31,19.6) {eat};

  \node[core, minimum width=17mm, minimum height=7mm] at (52,24.5) (panta) {\textbf{\panta-v2}};
  \draw[inner] (q2.east) -- (panta.west);

  \node[neut, minimum width=12mm, minimum height=7mm, font=\tiny] at (73,24.5) (inst)
    {\textbf{20k}\\instruct pairs};
  \draw[flow] (inst.west) -- (panta.east);
  \node[side, anchor=south] at (63.75,25.2) {tunes};

  \node[rowlab] at (17,9.5) {Answer};
  \node[chip] at (23,9.5) (a1) {};
  \icoClock{a1.center}{90}{90}
  \foreach \i/\h in {0/0.6, 1/1.4, 2/0.9, 3/2.1, 4/1.2, 5/2.4, 6/1.0, 7/1.7, 8/0.6} {
    \draw[speechLine, line width=0.55pt]
      ({33.5+1.3*\i},{9.5-\h}) -- ({33.5+1.3*\i},{9.5+\h});
  }
  \node[font=\scriptsize\itshape, text=neutText] at (52,9.5) (atxt) {``At noon.''};
  \node[pcap] at (23,4.6) {noon};
  \node[pcap, text=speechText] at (38.7,4.6) {speech};
  \node[pcap, text=neutText] at (52,4.6) {text};

  \draw[coreLine!80, line width=0.45pt] (panta.south) -- (52,16.5);
  \draw[coreLine!80, line width=0.45pt] (23,16.5) -- (52,16.5);
  \draw[inner] (23,16.5) -- (a1.north);
  \draw[inner] (38.7,16.5) -- (38.7,12.6);
  \draw[inner] (52,16.5) -- (52,11.4);

  \end{tikzpicture}%
  }
  \caption{Overview of our work. (a) RAPS recovers a concept missed by the
  rule-based grammar. (b) RAPS generates
  a pre-training corpus of 1.4M text-pictogram pairs used to train \textit{\panta-v1}. (c) \textit{\panta-v2}, initialized from \textit{\panta-v1} and instruction-tuned on 20k question-answer
  pairs}
\label{fig:overview-panta}
\end{figure}
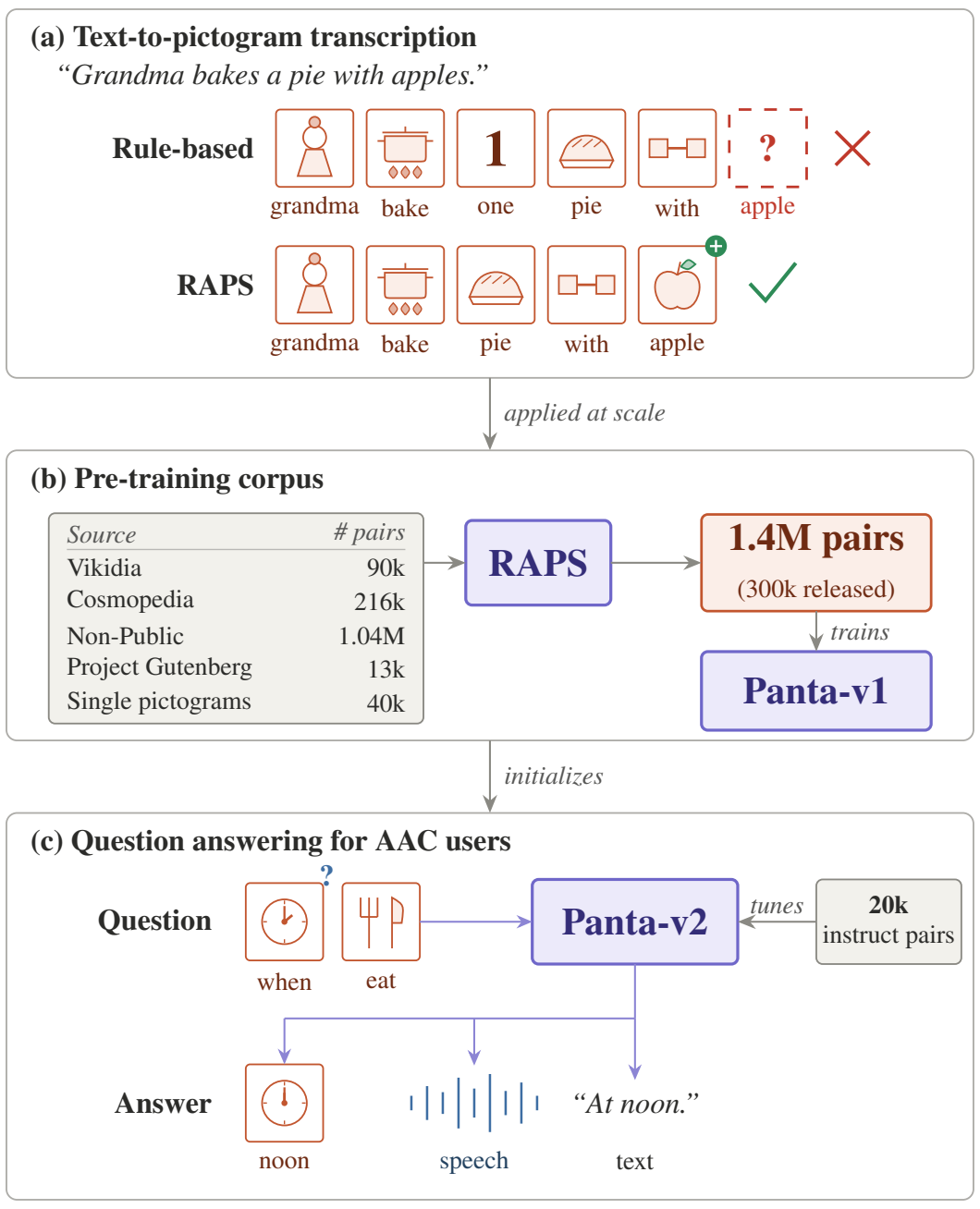

The main obstacle is the lack of training data.
Unlike text and speech corpora, which are abundant online, no large collection of naturally produced pictogram communication exists, to our knowledge \cite{dudy-bedrick-2018-compositional}.
Nor is there a widely adopted pictogram grammar: the same message can be expressed by several valid pictogram sequences, depending on the user and the context (Section~\ref{sec:transcription}).
Each text-to-pictogram system therefore defines its own hand-written rules \cite{vandeghinste2017translating, norre2021extending, dataset_cecile}, which depend on a fixed lexicon and that struggle with polysemous words \cite{sevens2016improving, dataset_cecile}.
As a result, in French, the only large-scale aligned corpus was generated by such rules, with only a small subset post-edited by experts \cite{dataset_cecile}.
Evaluating pictogram generation against these references largely measures agreement with the rule-based grammar rather than communicative adequacy, and neural systems trained on them inherit the limitations of the pipeline they are meant to replace.
Evidence from Spanish points in the same direction: on PICTOEDUCA \cite{pictoeduca}, a rule-based baseline still outperforms both a T5 model \cite{T5} fine-tuned on that baseline's output and a few-shot Llama-3.1-8B \cite{llama}.

We address data scarcity with a single annotation pipeline and a model trained on its output. The contributions of this work are as follows.
\begin{enumerate}
    \item We propose Retrieval-Augmented Pictogram Selection(RAPS), post-translation method that refines the output of a rule-based system through LLM-based concept extraction and retrieval over the ARASAAC\footnote{\url{https://arasaac.org/about-us}} vocabulary. Applied to French, RAPS yields a synthetic corpus of 1.4 million parallel pairs, of which we make 300k publicly released\footnote{The rest is withheld until approved by our legal department}.

    \item We release \textit{\panta}, the first instruction-tuned MLLM that both reads and generates pictogram sequences, serving communication support: it accepts pictogram or text input and answers in pictograms, speech, or text.

    \item We conduct a human preference study comparing human, rule-based, RAPS and model-generated pictogram sequences (660 rankings over 121 sentences), in which RAPS and our model significantly outperform the rule-based state of the art. Using these rankings, we show that standard reference-based metrics separate systems of clearly different quality but fail to rank strong systems against each other.

    \item We release an evaluation benchmark for French pictographic communication, comprising 121 human-ranked sentences with four transcriptions each.
\end{enumerate}

\section{Related Work}
\subsection{Multimodal Models}

Deep models were long built around a single modality, such as text \cite{bert, roberta}, images \cite{vit, resnet}, or audio \cite{wav2vec, hubert}.
Multimodal models instead take one or more modalities as input and may produce different output modalities.
A first class of approaches relies on contrastive learning, aligning representations of different modalities in a shared latent space by bringing together representations with the same semantic content and pushing apart those without \cite{clip, siglip, siglip2}.
Jointly training on multiple input modalities can also act as a regularizer and improve performance across all of them \cite{pantagruel}.
Generative multimodal models \cite{qwentwo, mini-omni} go further by producing several modalities as output, supporting richer human--machine interaction \cite{multi-modal-better}.

\subsection{Generative Multimodal Models}
Our work focuses on LLM-driven multimodal architectures. This overview is by no means exhaustive: we only describe the models that are closest to our design. 
LLaVA \cite{LLava} is an end-to-end model that processes images and text as input and generates textual outputs. 
Visual features are extracted by a pre-trained Vision Transformer (ViT) \cite{vit}, projected into the language model embedding space, and concatenated with text token embeddings before being fed to an LLM backbone. Ministral~3 \cite{ministral3} follows the same design, coupling a
$\sim$0.4B-parameter vision encoder with a multimodal projector and a
language model, and is released in 3B, 8B and 14B sizes under an Apache~2.0
license.

Mini-Omni \cite{mini-omni} extends this paradigm to speech generation by adding parallel output heads on top of the LLM, producing text and speech tokens simultaneously. 
Its successor, Mini-Omni2 \cite{mini-omni2}, adds visual inputs through a CLIP \cite{clip} encoder. 
More recently, Qwen Omni \cite{qwen-omni-3} introduces the \emph{Thinker-Talker} architecture, where the LLM backbone generates text while additional modality decoders take the LLM's hidden representations as input to produce outputs in other modalities.

\subsection{AAC and Pictogram Translation}
\vspace{-0.24cm}

Researchers have addressed AAC using deep learning in several ways. 
Neural work on AAC first focused on next-pictogram prediction, suggesting the next symbol in a sentence under construction \cite{pictobert, perreira}.
Text-to-pictogram translation is a widely studied task
\cite{macaire2025overview}, with rule-based systems for Dutch, English, Spanish, and French \cite{vandeghinste2017translating, sevens2015extending,
sevens2018words, norre2021extending, pictoeduca, dataset_cecile}.
The reverse direction, generating text from pictograms, has received less attention \cite{sevens-etal-2015-natural}.
Rule-based systems rely on linguistic pipelines combining tokenization, lemmatization, part-of-speech tagging or parsing, named entity recognition, and in some cases word sense disambiguation, and remain strong baselines (Section~\ref{sec:intro}).
In the ImageCLEFtoPicto shared task \cite{macaire2025overview}, a fine-tuned T5-large \cite{T5} obtained the best results, but it was trained on rule-generated data and evaluated against rule-based annotations \cite{Hjuler2025-pr}, so its scores also mainly reflect agreement with the pre-defined grammar.
While it is outside the scope of this paper, it is worth mentioning speech-to-pictogram translation that has been addressed with both cascaded and end-to-end models \cite{macaire2024towards}.
Unlike our approach, these systems support only one direction of translation, rather than both pictogram reading and generation.

\definecolor{pictoFill}{HTML}{FBEEE8}
\definecolor{pictoLine}{HTML}{C2562F}
\definecolor{pictoText}{HTML}{6E2A12}
\definecolor{coreFill}{HTML}{ECEAFA}
\definecolor{coreLine}{HTML}{6F67C9}
\definecolor{coreText}{HTML}{342D78}
\definecolor{neutFill}{HTML}{F2F1EC}
\definecolor{neutLine}{HTML}{8A8983}
\definecolor{neutText}{HTML}{2B2B29}

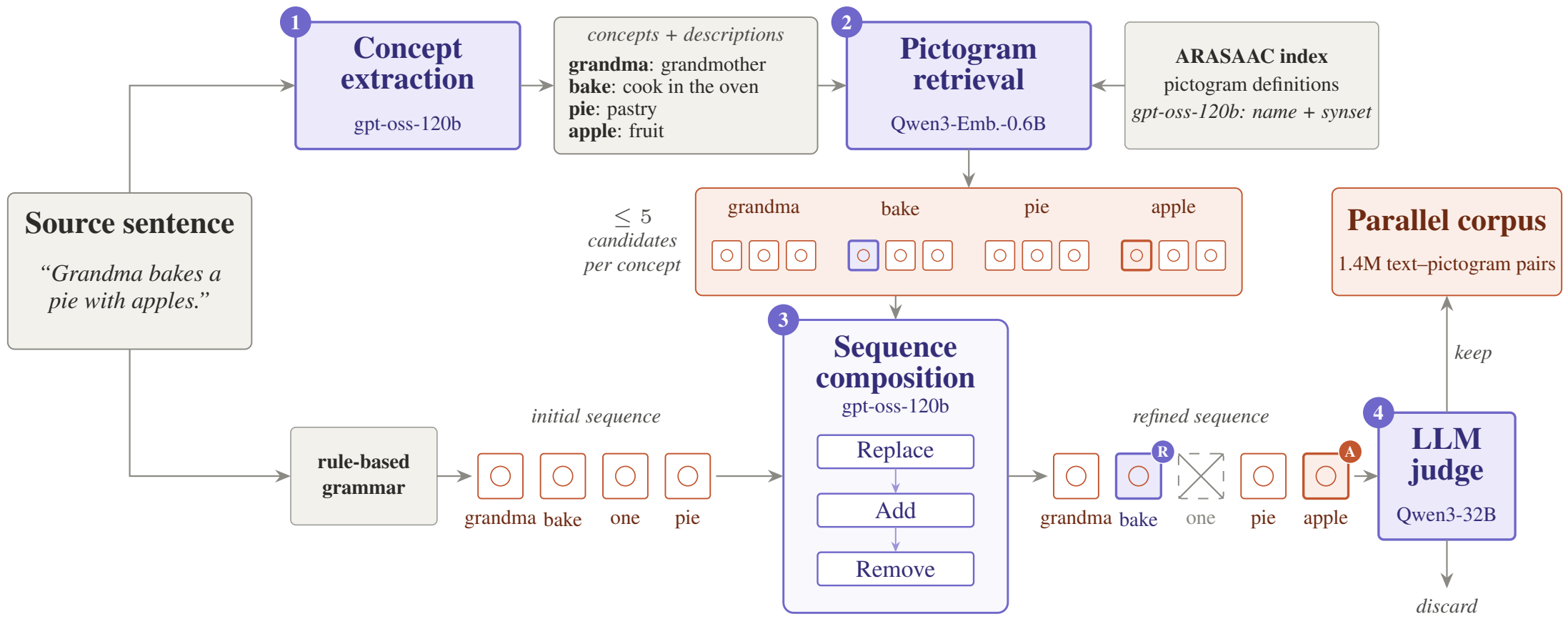
\begin{figure*}[th!]
  \centering
  \resizebox{0.9\textwidth}{!}{%
  \begin{tikzpicture}[
    x=1mm, y=1mm,
    font=\footnotesize,
    every node/.style={inner sep=0pt, outer sep=0pt},
    modality/.style={
      draw=neutLine, line width=0.4pt, fill=neutFill, rounded corners=2pt,
      text=neutText, align=center, inner sep=1.6pt},
    picto/.style={modality, draw=pictoLine, fill=pictoFill, text=pictoText},
    stage/.style={                       
      draw=coreLine, line width=0.7pt, fill=coreFill, rounded corners=1.5pt,
      text=coreText, align=center, inner sep=1.5pt},
    substage/.style={
      draw=coreLine, line width=0.5pt, fill=white, rounded corners=1pt,
      text=coreText, align=center, inner sep=0.8pt, font=\scriptsize,
      minimum width=16mm, minimum height=3.4mm},
    groupbox/.style={
      draw=coreLine, line width=0.6pt, fill=coreFill!35, rounded corners=3pt},
    detok/.style={
      draw=neutLine!80, line width=0.35pt, fill=neutFill, rounded corners=1.5pt,
      text=neutText, align=center, inner sep=1.2pt, font=\tiny},
    badge/.style={
      circle, fill=coreLine, text=white, font=\tiny\bfseries,
      minimum size=3.2mm},
    opbadge/.style={
      circle, draw=white, line width=0.3pt, text=white,
      font=\fontsize{4.5}{4.5}\selectfont\bfseries, minimum size=2.4mm},
    chip/.style={
      draw=pictoLine, line width=0.4pt, fill=white, rounded corners=1pt,
      minimum width=4.6mm, minimum height=4.6mm},
    minichip/.style={
      draw=pictoLine, line width=0.35pt, fill=white, rounded corners=0.8pt,
      minimum width=3.0mm, minimum height=3.0mm},
    replaced/.style={draw=coreLine, fill=coreFill, line width=0.7pt},
    added/.style={draw=pictoLine, fill=pictoFill, line width=0.7pt},
    removed/.style={draw=neutLine, dashed, fill=white},
    pictocap/.style={font=\tiny, text=pictoText, align=center},
    flow/.style={-{Stealth[length=1.6mm, width=1.3mm]},
                 draw=neutLine, line width=0.45pt},
    inner/.style={-{Stealth[length=1.1mm, width=0.9mm]},
                  draw=coreLine!70, line width=0.4pt},
    note/.style={font=\tiny\itshape, text=neutText!85, align=center},
  ]

  \node[modality, minimum width=25mm, minimum height=16mm] at (13.5,-28) (src) {};
  \node[text=neutText] at (13.5,-23.0) {\textbf{Source sentence}};
  \node[text=neutText, font=\scriptsize\itshape, align=center] at (13.5,-29.8)
    {``Grandma bakes a\\pie with apples.''};

  \node[stage, minimum width=23mm, minimum height=13mm] at (42,-9) (s1)
    {\textbf{Concept}\\[-1pt]\textbf{extraction}\\[2pt]{\tiny gpt-oss-120b}};
  \node[badge] at (s1.north west) {1};

  \node[modality, minimum width=27mm, minimum height=14mm] at (70.5,-9) (concepts) {};
  \node[note] at (70.5,-3.9) {concepts + descriptions};
  \node[font=\tiny, text=neutText, align=left, anchor=north west] at (58.5,-6.0)
    {\textbf{grandma}: grandmother\\[-0.5pt]
     \textbf{bake}: cook in the oven\\[-0.5pt]
     \textbf{pie}: pastry\\[-0.5pt]
     \textbf{apple}: fruit};

  \node[stage, minimum width=25mm, minimum height=13mm] at (99.5,-9) (s2)
    {\textbf{Pictogram}\\[-1pt]\textbf{retrieval}\\[2pt]{\tiny Qwen3-Emb.-0.6B}};
  \node[badge] at (s2.north west) {2};

  \node[detok, minimum width=26mm, minimum height=13mm] at (128.5,-9) (index)
    {\textbf{ARASAAC index}\\[1pt]pictogram definitions\\[1pt]
     {\itshape gpt-oss-120b: name + synset}};

  \draw[flow] (src.north) |- (s1.west);
  \draw[flow] (s1.east) -- (concepts.west);
  \draw[flow] (concepts.east) -- (s2.west);
  \draw[flow] (index.west) -- (s2.east);

  \node[picto, minimum width=56mm, minimum height=11mm] at (99.5,-25) (pool) {};
  \node[note, anchor=east] at (70,-25) {$\leq 5$\\candidates\\per concept};
  \foreach \gx/\word/\sel in {78.5/grandma/0, 92.5/bake/1, 106.5/pie/0, 120.5/apple/2} {
    \node[pictocap] at (\gx,-21.6) {\word};
    \foreach \k in {-1,0,1} {
      \ifnum\sel=0
        \node[minichip] at ({\gx+3.8*\k},-26.4) (c) {};
      \else
        \ifnum\k=-1
          \ifnum\sel=1
            \node[minichip, replaced] at ({\gx+3.8*\k},-26.4) (c) {};
          \else
            \node[minichip, added] at ({\gx+3.8*\k},-26.4) (c) {};
          \fi
        \else
          \node[minichip] at ({\gx+3.8*\k},-26.4) (c) {};
        \fi
      \fi
      \draw[pictoLine, line width=0.3pt] (c.center) circle (0.6mm);
    }
  }
  \draw[flow] (s2.south) -- (pool.north -| s2.south);

  \node[detok, minimum width=15mm, minimum height=10mm] at (37.5,-49) (gram)
    {\textbf{rule-based}\\[0.5pt]\textbf{grammar}};
  \draw[flow] (src.south) |- (gram.west);

  \node[note] at (61.3,-43) {initial sequence};
  \foreach \x/\word in {51.5/grandma, 57.9/bake, 64.3/one, 70.7/pie} {
    \node[chip] at (\x,-49) (c) {};
    \draw[pictoLine, line width=0.35pt] (c.center) circle (1.0mm);
    \node[pictocap] at (\x,-53.4) {\word};
  }
  \draw[flow] (gram.east) -- (48.6,-49);

  \node[groupbox, minimum width=23mm, minimum height=30mm] at (92,-48) (s3) {};
  \node[badge] at (s3.north west) {3};
  \node[text=coreText, align=center] at (92,-37.6)
    {\textbf{Sequence}\\[-1pt]\textbf{composition}};
  \node[text=coreText, font=\tiny] at (92,-42.0) {gpt-oss-120b};
  \node[substage] at (92,-46.5) (rep) {Replace};
  \node[substage] at (92,-52.5) (add) {Add};
  \node[substage] at (92,-58.5) (rem) {Remove};
  \draw[inner] (rep.south) -- (add.north);
  \draw[inner] (add.south) -- (rem.north);

  \draw[flow] (73.6,-49) -- (s3.west |- 0,-49);
  \draw[flow] (92,-30.5) -- (s3.north);

  \node[note] at (123.3,-43) {refined sequence};
  \node[chip] at (110.5,-49) (f1) {};
  \node[chip, replaced] at (116.9,-49) (f2) {};
  \node[chip, removed] at (123.3,-49) (f3) {};
  \node[chip] at (129.7,-49) (f4) {};
  \node[chip, added] at (136.1,-49) (f5) {};
  \foreach \n in {f1,f2,f4,f5} {
    \draw[pictoLine, line width=0.35pt] (\n.center) circle (1.0mm);
  }
  \draw[neutLine, line width=0.45pt] (f3.north west) -- (f3.south east);
  \draw[neutLine, line width=0.45pt] (f3.south west) -- (f3.north east);
  \node[opbadge, fill=coreLine]  at ([shift={(0.2,0.2)}]f2.north east) {R};
  \node[opbadge, fill=pictoLine] at ([shift={(0.2,0.2)}]f5.north east) {A};
  \foreach \x/\word/\col in {110.5/grandma/pictoText, 116.9/bake/coreText,
                             123.3/one/neutLine, 129.7/pie/pictoText,
                             136.1/apple/pictoText} {
    \node[pictocap, text=\col] at (\x,-53.4) {\word};
  }
  \draw[flow] (s3.east |- 0,-49) -- (107.6,-49);

  \node[stage, minimum width=14mm, minimum height=13mm] at (148.5,-49) (s4)
    {\textbf{LLM}\\[-1pt]\textbf{judge}\\[2pt]{\tiny Qwen3-32B}};
  \node[badge] at (s4.north west) {4};
  \draw[flow] (139,-49) -- (s4.west);

  \node[picto, minimum width=18mm, minimum height=11mm] at (148.5,-25) (corpus)
    {\textbf{Parallel corpus}\\[1.5pt]{\tiny 1.4M text--pictogram pairs}};
  \draw[flow] (s4.north) -- node[note, right=0.8mm] {keep} (corpus.south);
  \draw[flow] (s4.south) -- (148.5,-60.5);
  \node[note] at (148.5,-62.3) {discard};

  \end{tikzpicture}%
  }
  \caption{Overview of Retrieval-Augmented Pictogram Selection.}
  \label{fig:raps}
\end{figure*}


\section{Method}

We introduce Retrieval-Augmented Pictogram Selection (RAPS), a new post-translation method that refines the output of a rule-based system by correcting translation errors without adding new rules. We use RAPS to generate a large synthetic parallel text–pictogram corpus, which we then leverage to fine-tune two models, including an instruction-tuned model, for multimodal communication. These models take text and/or pictograms as input and generate text, pictograms, or speech. Training models both validates the quality of the corpus and shows how to distil the RAPS agentic process to make it faster.

\subsection{Text-Pictogram Dataset}

\subsubsection{Retrieval-Augmented Pictogram Selection}
RAPS updates the rule-based alignment in four steps (see Fig.~\ref{fig:raps}): an LLM extracts key concepts, a retriever finds candidate pictograms, an LLM updates the input sequence with the retrieved pictograms, and an LLM judge filters out pairs missing a core concept (Full prompts in section~\ref{appendix:prompts}).
Each ARASAAC pictogram is enriched with two descriptions: a semantic description, a textual definition generated by an LLM from the pictogram name and, when available, its WordNet synset; and a visual description, generated by a VLM from the pictogram image.

\paragraph{Stage 1: Concept extraction.}
Given a source sentence, an LLM lists its \textit{concept candidates}, the words or groups of words that carry its core meaning, each with a short definition that resolves polysemy.
The prompt also specifies how to handle the cases where rule-based systems fail: named entities are replaced by hypernyms and salient properties, complex concepts are split into simpler parts, and negations are made explicit.

\paragraph{Stage 2: Pictogram retrieval.}
For each concept candidate produced in Stage 1, a retrieval module embeds its description and the semantic descriptions of all ARASAAC pictograms into a shared space, and retrieves up to five pictograms with the highest cosine similarity.

\paragraph{Stage 3: Sequence composition.}
A text-only LLM edits the output of the rule-based system \cite{dataset_cecile} using the Stage~2 candidates, each represented by its lemma, its semantic description and its visual description.
Stage~3 performs three operations:  

\textbf{Replace:} exchange a pictogram for a better candidate when one is found. The LLM is prompted to judge candidates by semantic description first, then by visual description.

\textbf{Add:} complement with pictograms from the candidates when a core concept from the input is missing.

\textbf{Remove:} remove pictograms that are unnecessary for understanding the sentence.

\paragraph{Stage 4: Quality filtering.}
As a final validation step, an independent LLM judges each text-pictogram pair by representing pictograms with their associated lemma and discarding pairs where the pictogram sequence does not cover a core concept of the source sentence.

Although applied here to French, RAPS is not specific to it; porting it to another language requires a pictogram lexicon, a text-to-pictogram translation system, such as the rule-based ones available for Dutch, English or Spanish, and adapting the language-specific rules and examples of its prompts. If no baseline translation system is available, directly prompting the text LLM to compose a pictogram sequence from the retrieved candidates is a viable strategy.

\subsection{Training an LLM in the pictogram domain}

To validate the quality of the RAPS approach, we train \textit{\panta} on synthetic corpora created by translating text to pictograms. 
Our \textit{Thinker-Talker} architecture \cite{qwen-omni-3} builds on a pretrained VLLM that generates textual representations. Additional neural network layers use the hidden representations of the text-generating component to produce outputs in other modalities. More details on \textit{\panta}'s architecture can be found in Appendix~ \ref{appendix:model-details}.
One challenge is to keep the instruct capabilities of the LLM with limited finetuning. 
We represent pictograms as visual tokens generated by the vision encoder of \textit{\panta}, downsampled to 4 tokens per image for efficiency purposes. Using a vision encoder, compared to embedding pictogram identifiers, allows us to benefit from the vision prior and potentially input pictogram variants not seen in the lexicon.
Modality-specific generation is handled by lightweight shallow decoders operating in an early-exit fashion: at a selected intermediate layer of the backbone, the hidden states are branched off and fed into dedicated decoders, one per output modality, while the main backbone continues processing for text generation.
Each modal decoder operates autoregressively, where the task is reduced to token prediction.
The pictogram decoder predicts, at each step, the identifier of a pictogram from the ARASAAC lexicon.
For speech generation, we follow the approach described in \cite{sparktts}: the decoder was trained to predict the discrete semantic tokens of the BiCodec tokenizer vocabulary \cite{sparktts}, which can be converted to a waveform by the BiCodec decoder. While this model has speech capabilities, testing them is not in scope of this paper.

\subsubsection{Pre-training Dataset}

We make a key assumption to guide our data collection strategy: pictogram sequences are primarily used to convey simple, short, unambiguous sentences.
This observation motivated us to focus on sources that naturally produce this type of language, namely content written for children and teenagers.
Hence, our dataset was constructed from the following four sources:
(1) children's literature extracted from Project Gutenberg~\cite{gutenberg}, a digital library of public-domain works, which adds narrative and literary styles; 
(2) Vikidia~\cite{vikidia}, a French encyclopedia for readers aged 8 to 13, which offers simplified explanations of a wide range of topics;
(3) Cosmopedia~\cite{cosmopedia}, a large collection of AI-generated educational content, including textbooks, blog posts, stories and how-to articles, part of which targets younger audiences; and
(4) \anonymousdataset{}, a collection of instructional web articles on everyday topics.
The extraction pipeline is provided in Appendix \ref{appendix:data-preprocessing-raps}, and Table \ref{tab:pictodataset} reports the corpus statistics.
\begin{table}[b]
  \centering
  \small
  \begin{tabular}{lr}
    \hline
    \textbf{Dataset} & \textbf{\# Pairs} \\
    \hline
    Gutenberg \cite{gutenberg}  & $12\,707$     \\
    Vikidia \cite{vikidia}              & $90\,431$     \\
    Cosmopedia \cite{cosmopedia}        & $215\,666$    \\
    \anonymousdataset{}                 & $1\,038\,004$ \\
    Single picto                        & $40\,000$     \\
    \hline
    \textit{Total}                      & $1\,396\,808$ \\
    \hline
    Propicto$^{*}$ \cite{dataset_cecile} & $528\,000$ \\
    \hline
  \end{tabular}
  \caption{\label{tab:pictodataset}
    Number of text-pictogram pairs per source, after filtering.
    $^{*}$Used for pictogram-to-text only.}
\end{table}

\subsubsection{Instruction Dataset}
\label{sec:instruct_dataset}

We also built a French instruction dataset of question-answer pairs. 
Questions are short and self-contained, and answers follow the French Easy-to-Read guidelines (FALC): a single sentence with concrete vocabulary that can be transcribed into pictograms. 
The training set contains $19\,494$ pairs: $14\,965$ synthetic questions generated from everyday situations and $4\,529$ questions extracted from \anonymousdataset{}. The test set contains $1\,285$ pairs from French-Instruct\footnote{\url{https://huggingface.co/datasets/angeluriot/french_instruct}}.
We then convert each answer into a pictogram sequence with RAPS and into speech with a text-to-speech model. Appendix~\ref{appendix:instruct_dataset} details the dataset construction.

\section{Experimental Details}
Semantic descriptions are generated with \textit{gpt-oss-120b}~\cite{gptoss120b} and visual descriptions with \textit{Qwen3-VL-8B-Instruct}~\cite{Qwen3-vl}.
For RAPS, we use \textit{openai/gpt-oss-120b}~\cite{gptoss120b} at every stage, \textit{Qwen3-Embedding-0.6B}~\cite{qwen3embedding} for the text embeddings of Stage~2, and \textit{Qwen3-32B}~\cite{qwen3} for the filtering stage.
For the Instruction Dataset, \textit{Qwen3-32B}~\cite{qwen3} generates the answers to the questions, RAPS provides the text-to-pictogram annotations, and \textit{Voxtral-4B-TTS-2603}~\cite{voxtral} %
produces the speech.
As a backbone, we adopt \textit{Ministral-3-8B-Instruct-2512}~\cite{ministral3}, a multilingual, vision-capable instruction-tuned model released by Mistral AI.
We train \textit{\panta} in two stages, following standard practice~\cite{mini-omni}. The first stage, \textit{modality alignment}, trains \textit{\panta} on pictogram transcription, text-to-pictogram, and text-to-speech, yielding \textit{\panta-v1}. The second stage, \textit{multitask instruction tuning}, continues training on these tasks alongside our Instruction Dataset \ref{sec:instruct_dataset}, yielding \textit{\panta-v2}. Further implementation details are given in Appendix~\ref{appendix:implementation-details}.
\section{Results}
\label{sec:results}


\subsection{Manual text-to-Pictograms evaluation} \label{sec:transcription}
\begin{table}[b]
  \centering
  \small
  \setlength{\tabcolsep}{5pt}
  \begin{tabular}{@{}lccc@{}}
    \hline
    \textbf{System} & \textbf{Rank} $\downarrow$ & \textbf{Win} $\uparrow$ & \textbf{Loss} $\downarrow$ \\
    \hline
    Human        & $2.48 \pm 1.23$          & $\mathbf{32.4\%}$ & $30.9\%$          \\
    PictoGrammar & $3.00 \pm 1.05$          & $13.0\%$          & $42.4\%$          \\
    RAPS         & $2.34 \pm 1.01$          & $24.4\%$          & $15.6\%$          \\
    \panta-v1     & $\mathbf{2.18} \pm 0.99$ & $30.2\%$          & $\mathbf{11.1\%}$ \\
    \hline
  \end{tabular}
  \caption{\label{tab:ranking}
    Human ranking of the text-to-pictogram systems.
    Rank: mean rank $\pm$ standard deviation. Win/Loss: percentage of first/last places.}
\end{table}

We compare four systems in a human ranking experiment (interface in Appendix~\ref{appendix:interface}): (1) \textbf{Human}, pictogram sequences produced
manually by trained annotators, none of whom are authors of this work;
(2) \textbf{PictoGrammar}, the French symbolic pipeline of
\citet{dataset_cecile}; (3) \textbf{RAPS}, our Retrieval-Augmented Pictogram
Selection approach, applied without its quality-filtering step; and
(4) \textbf{\panta-v1}, our model trained on the RAPS-generated dataset.
Human sequences come from four French corpora annotated with ARASAAC pictograms: \textit{Medical}~\cite{norre2025thesis,bouillon2017babeldr}, \textit{Email}~\cite{norre2021extending,sevens2017simplified}, \textit{Book}~\cite{vaschalde2018automatic,norre2021extending} and \textit{Arasaac materials} (details in Appendix~\ref{appendix:ground-true-dataset}). We consider \textit{Medical} the most reliable reference: it consists of short doctor-to-patient questions written directly in French and partly transcribed by a speech therapist, whereas \textit{Email} was translated from Dutch and \textit{Book} contains longer, more varied sentences.
Participants were asked to rank four anonymous translations of a French sentence, shown in random order, following the instruction: \textit{``Rank these translations from the one that best conveys
the message to the one that least conveys it, according to your own judgment.''}.
We chose a ranking design over Likert scales because ranking elicits more reliable judgments and is less prone to order effects~\cite{rating}.
Sixty-three participants from our laboratory, aged between 18 and 70, took part. Each participant answered twelve questions, and ten participants dropped out before completing their rankings. In total, we collected 660 individual rankings.

Our corpus comprises 58 Medical, 20 Email, 20 Book, and 23 Arasaac sequences, making 121 ranked sequences across 4 models, with moderate inter-annotator agreement (mean Kendall's $W = 0.43$ computed per
sentence and averaged). 
\begin{table}[b]
  \centering
  \small
  \setlength{\tabcolsep}{5pt}
  \begin{tabular}{@{}lccccc@{}}
    \hline
    \textbf{System} & \textbf{Medical} & \textbf{Email} & \textbf{Book} & \textbf{AM} & \textbf{All} \\
    \hline
    Human        & $\mathbf{2.05}$ & $3.11$          & $2.89$          & $2.65$          & $2.48$          \\
    PictoGrammar & $3.28$          & $2.68$          & $2.62$          & $2.92$          & $3.00$          \\
    RAPS         & $2.35$          & $2.39$          & $2.36$          & $2.24$          & $2.34$          \\
    \panta-v1     & $2.32$          & $\mathbf{1.83}$ & $\mathbf{2.12}$ & $\mathbf{2.19}$ & $\mathbf{2.18}$ \\
    \hline
  \end{tabular}
  \caption{\label{tab:perdataset}
    Mean rank per source corpus (lower is better). AM: Arasaac Material.}
\end{table}

Table~\ref{tab:ranking} provides the results of the evaluation. \textit{\panta-v1} obtains the best mean rank (2.18), followed by RAPS (2.34), the human sequences (2.48), and PictoGrammar (3) (Table~\ref{tab:ranking}). RAPS, \textit{\panta-v1} and the human references are all ranked significantly above PictoGrammar: on per-sentence mean ranks ($n = 121$), a Friedman test rejects the hypothesis that all systems are ranked equally ($p < 0.05$), and pairwise Wilcoxon signed-rank tests with Bonferroni correction confirm that each of the three systems outperforms PictoGrammar, but the three other approaches are not statistically different.
The human references (Table~\ref{tab:perdataset}) are the most often ranked first (32.4\%) but also often last (30.9\%), reflecting the uneven quality of the source corpora: they rank best on \textit{Medical}, our most reliable reference (2.05), but last on \textit{Email} and \textit{Book} datasets.

\begin{table*}[b!th]
  \centering
  \small
  \setlength{\tabcolsep}{4pt}
  \resizebox{0.9\textwidth}{!}{%
  \begin{tabular}{l cccc cccc}
    \hline
    & \multicolumn{4}{c}{\textbf{All sentences (121)}} & \multicolumn{4}{c}{\textbf{Medical only (58)}} \\
    \cline{2-5} \cline{6-9}
    \textbf{Metric} & PG/\panta-v1 & PG/RAPS & RAPS/\panta-v1 & Mean & PG/\panta-v1 & PG/RAPS & RAPS/\panta-v1 & Mean \\
    \hline
    BLEU (lemmas)   & $0.333^{*}$ & $0.333^{*}$ & $0.172$ & $0.279$ & $0.720^{*}$ & $0.608^{*}$ & $\mathbf{0.306}$ & $0.545$ \\
    METEOR (lemmas) & $0.156$ & $0.321^{*}$ & $0.140$ & $0.206$ & $0.480^{*}$ & $0.608^{*}$ & $0.184$ & $0.424$ \\
    WER (lemmas)    & $\mathbf{0.528}^{*}$ & $\mathbf{0.500}^{*}$ & $0.038$ & $\mathbf{0.355}$ & $\mathbf{0.808}^{*}$ & $\mathbf{0.755}^{*}$ & $0.105$ & $0.556$ \\
    chrF++ (lemmas) & $0.162$ & $0.197$ & $0.097$ & $0.152$ & $0.500^{*}$ & $0.491^{*}$ & $0.179$ & $0.390$ \\
    \hline
    BLEU (IDs)      & $0.200$ & $0.333^{*}$ & $0.149$ & $0.227$ & $0.619^{*}$ & $0.644^{*}$ & $0.167$ & $0.477$ \\
    METEOR (IDs)    & $0.060$ & $0.208$ & $\mathbf{0.191}$ & $0.153$ & $0.381$ & $0.435$ & $0.167$ & $0.327$ \\
    WER (IDs)       & $0.443^{*}$ & $0.455^{*}$ & $0.096$ & $0.331$ & $0.760^{*}$ & $0.750^{*}$ & $0.220$ & $\mathbf{0.577}$ \\
    chrF++ (IDs)    & $-0.162$ & $-0.060$ & $0.062$ & $-0.053$ & $0.000$ & $0.055$ & $0.053$ & $0.036$ \\
    \hline
    Mean            & $0.215$ & $0.286$ & $0.118$ & $0.206$ & $0.533$ & $0.543$ & $0.172$ & $0.416$ \\
    \hline
    Human (LOO)     & $0.491$ & $0.357$ & $0.317$ & $0.388$ & $0.520$ & $0.382$ & $0.371$ & $0.424$ \\
    \hline
  \end{tabular}%
  }
  \caption{Agreement ($\tau$) between metric and human preferences for each
    pair of systems (PG = PictoGrammar), using the human sequences as
    references. $^{*}$: significant agreement after
    Holm correction ($p < 0.05$).}
  \label{tab:metrics}
\end{table*}

\subsection{Reliability of automatic metrics}
\label{sec:metrics}
Many works \cite{norre2021extending, pictoeduca} suggest that current automatic metrics are not reliable enough to evaluate text-to-pictogram generation. 
Moreover, prior works made evaluations at different levels. For instance,  ImageCLEFtoPicto \cite{macaire2025overview} compares ARASAAC lemmas, ignoring visual content even though many pictograms share the same lemma, whereas others compare pictogram identifiers \cite{pictoeduca}, which disregards similarity between pictograms.
We test whether reference-based metrics agree with the human rankings of Sec.~\ref{sec:transcription}, using the human sequences as reference. We consider four metrics, computed at the segment level on both ARASAAC lemmas and identifiers: (1) \textbf{BLEU}~\cite{papineni2002bleu}, n-gram precision with a brevity
penalty; (2) \textbf{METEOR}~\cite{banerjee2005meteor}, unigram alignment with
stem and synonym matching; (3) \textbf{WER}, edit distance normalized by the
reference length; and (4) \textbf{chrF++}~\cite{popovic2017chrf}, a character
and word n-gram F-score.

For each pair of systems, we compare the preference induced by the metric (the system with the better score) with the preference of the annotators (the system with the better mean rank), sentence by sentence, discarding ties on either side.
We report the agreement as $\tau = 2 \times \text{accuracy} - 1$, which equals Kendall's $\tau_b$ when only two systems are compared.
Results are reported on two corpora, all sentences and the Medical subset of the human evaluation.
We test whether each $\tau$ differs from 0 with an exact two-sided binomial test, and apply a Holm correction separately within each corpus over its 24 tests.
For the row Human (LOO), we compute the same pairwise $\tau$ in a leave-one-out fashion: for each pair of systems, the preference of one annotator is compared with the preference given by the mean rank of the remaining annotators of the same sentence.
This value serves as a human reference for the level a metric can be expected to reach. Table~\ref{tab:metrics} shows that the metrics separate systems of clearly different quality, but not systems of similar quality.
WER obtains the highest mean $\tau$ on both corpora and for both representations, but this advantage comes from the pairs involving PictoGrammar, where the quality gap is large.
The RAPS/\textit{\panta-v1} pair is the hardest to separate: it opposes the two best systems, and their outputs are close since \textit{\panta-v1} is trained on data generated by RAPS.
On this pair, no metric remains significant after correction, neither on all sentences nor on the Medical subset.
This failure is not due to a lack of human consensus: on all sentences, annotators agree with each other to a similar extent on the PictoGrammar/RAPS and RAPS/\textit{\panta-v1} pairs ($\tau = 0.36$ and $0.32$), yet no metric approaches this level on RAPS/\textit{\panta-v1}. Regarding the representation, BLEU and WER give similar results on lemmas and on identifiers, whereas METEOR and chrF++ degrade on identifiers.
On identifiers, METEOR can no longer use stems and synonyms and reduces to exact matching, while chrF++ compares the digits of the identifiers. These metrics therefore cannot replace human judgments to compare close systems, such as RAPS variants in an ablation study, left to future work.

\subsection{Pictogram-to-Text Translation}
\label{sec:p2t-eval}
\begin{table}[t]
  \centering
  \small
  \setlength{\tabcolsep}{4pt}
  \resizebox{0.8\columnwidth}{!}{%
  \begin{tabular}{lcccccc}
    \hline
    & \multicolumn{3}{c}{\textbf{Email}} & \multicolumn{3}{c}{\textbf{Medical}} \\
    \cline{2-4} \cline{5-7}
    \textbf{$k$} & \textbf{BS}$\uparrow$ & \textbf{BLEU}$\uparrow$ & \textbf{WER}$\downarrow$ & \textbf{BS}$\uparrow$ & \textbf{BLEU}$\uparrow$ & \textbf{WER}$\downarrow$ \\
    \hline
    \multicolumn{7}{l}{\textit{Ministral-3-8B (lemmas)}} \\
    $0$ & $36.6$          & $\mathbf{17.3}$ & $91.6$          & $36.8$          & $11.1$          & $101.5$         \\
    $2$ & $\mathbf{43.5}$ & $\mathbf{22.5}$ & $\mathbf{74.3}$ & $41.4$          & $12.4$          & $95.6$          \\
    $4$ & $\mathbf{43.4}$ & $\mathbf{22.0}$ & $\mathbf{73.9}$ & $44.2$          & $15.2$          & $85.0$          \\
    $8$ & $45.1$          & $\mathbf{24.9}$ & $\mathbf{69.5}$ & $49.5$          & $22.1$          & $72.8$          \\
    \hline
    \multicolumn{7}{l}{\textit{\panta-v1 (IDs)}} \\
    $0$ & $27.5$          & $12.2$          & $91.4$          & $42.0$          & $17.7$          & $82.8$          \\
    $2$ & $32.5$          & $14.7$          & $85.0$          & $48.5$          & $23.3$          & $72.2$          \\
    $4$ & $34.3$          & $15.8$          & $82.9$          & $50.1$          & $24.4$          & $69.8$          \\
    $8$ & $35.9$          & $16.4$          & $82.0$          & $51.7$          & $26.3$          & $65.8$          \\
    \hline
    \multicolumn{7}{l}{\textit{\panta-v1}} \\
    $0$ & $\mathbf{37.3}$ & $15.3$          & $\mathbf{88.1}$ & $\mathbf{51.5}$ & $\mathbf{24.0}$ & $\mathbf{71.6}$ \\
    $2$ & $39.8$          & $17.3$          & $78.0$          & $\mathbf{54.7}$ & $\mathbf{26.4}$ & $\mathbf{65.3}$ \\
    $4$ & $42.5$          & $19.3$          & $74.9$          & $\mathbf{54.7}$ & $\mathbf{28.5}$ & $\mathbf{62.4}$ \\
    $8$ & $\mathbf{45.9}$ & $21.4$          & $71.2$          & $\mathbf{56.0}$ & $\mathbf{31.1}$ & $\mathbf{59.5}$ \\
    \hline
  \end{tabular}%
  }
  \caption{Pictogram-to-text translation results}
  \label{tab:p2t}
\end{table}

Pictogram-to-text translation is essential for communication between AAC and non-AAC speakers and, in our case, serves as an indirect measure of a system's ability to interpret pictogram input. 
Yet it has received far less attention than the reverse direction \citep{macaire2025overview,pictoeduca}. We evaluate it on two corpora used in Section: \ref{sec:results} \textit{Medical} \citep{norre2025thesis} and \textit{Email} \citep{norre2021extending,sevens2017simplified}, with 240 and 103 test sentences, respectively, after length filtering.

We compare three models, all with a frozen backbone, so that only the pictogram representation differs:
(1)~\textbf{Ministral-3-8B} prompted with each pictogram replaced by its first French ARASAAC lemma;
(2)~\textbf{\panta-v1 (IDs)}, where each pictogram identifier is mapped to a vector by a learned embedding matrix of size $|V| \times d$, with $V$ the ARASAAC vocabulary (same training as \textit{\panta-v1});
(3)~\textbf{\panta-v1}, where each pictogram image is encoded by the Pixtral vision encoder of the backbone, followed by attention pooling (Appendix~\ref{appendix:model-details}).
For $k \in \{0, 2, 4, 8\}$, the prompt includes $k$ few-shot examples sampled
from held-out pairs of the same corpus (disjoint from the test set), identical
for all models; scores are averaged over 5 seeds.
We report BERTScore \cite{bertscore} with CamemBERT \cite{camembert} (rescaled), BLEU and WER, computed after case and punctuation normalization. \textit{\panta}-v1 (IDs) is the weakest model on both corpora (Table~\ref{tab:p2t}). 
A likely explanation is that many pictograms are rare in the training data, so their embeddings receive few updates, whereas the image encoder shares visual features across pictograms and can, in principle, encode pictograms never seen during training, which matters as the ARASAAC vocabulary keeps growing. 
On \textit{Medical}, our most reliable reference and the use case closest to our target, \textit{\panta-v1} outperforms Ministral-3-8B on every metric and for every $k$, even though Ministral-3-8B receives the lemmas directly as text. 
As $k$ grows, the gap narrows because demonstrations teach Ministral-3-8B in context how to read a pictogram sequence, which \textit{\panta-v1} already learned during training.

On \textit{Email}, results are mixed: Ministral-3-8B is better on surface
metrics (BLEU, WER), while \textit{\panta-v1} is competitive on BERTScore, so
no system clearly outperforms the other. These references were translated from
Dutch, often with unnatural phrasing, and contain named entities with no
corresponding pictogram.
\subsection{Instruction-Following Results}
\label{sec:auto-eval}
The instruction following capability was evaluated on 1,285 held-out question-answer pairs from the French-Instruct part of our dataset (Section~\ref{sec:instruct_dataset}), which covers all input and output modalities.
Each question is given either as text or as a pictogram sequence, and the answer is created in all modalities: pictograms, speech, and text.
We compare \textit{\panta-v2} with its backbone, \textit{Ministral-3-8B} \cite{ministral3}, in 0-shot and 8-shot settings; since the backbone accepts only text, this comparison is restricted to text-to-text.
We report BLEU \cite{papineni2002bleu}, ROUGE-L \citep{rouge} and BERTScore \citep{bertscore} with Camembert \cite{camembert} re-scale against the reference answers.
Following the LLM-as-a-judge paradigm, \textit{Mistral-Small-3.2-24B} \footnote{https://huggingface.co/mistralai/Mistral-Small-3.2-24B-Instruct-2506}, \textit{gpt-oss-120b} \cite{gptoss120b} and \textit{deepseek-ai/DeepSeek-R1-Distill-Qwen-32B} \cite{deepseek} rate each answer for relevance, correctness and simplicity on a 5-point scale mapped to $[0,1]$, and the overall score is their mean. 
Prompts are given in Appendix~\ref{appendix:prompts-judge}.

Two authors of this paper evaluated 40 questions per setting (440 answers in
total; protocol in Appendix~\ref{app:human-instruct}). In both cross-modal
settings, text question with pictogram answer and pictogram question with text
answer, annotators rate two aspects: \textit{correctness}, whether the answer
of \textit{\panta-v2} correctly answers the question in the other modality,
and \textit{transcription quality}, whether the pictogram sequence preserves
the meaning of its textual counterpart (the text answer of \textit{\panta-v2},
or the original question transcribed by RAPS). To validate the LLM judges,
annotators also rate the text answers of \textit{\panta-v2},
\textit{Ministral-3-8B} (0-shot) and the reference answers for correctness and
simplicity, on the same 5-point scale as the judges.

\subsubsection{Text Output Evaluation}
\begin{table}[!h]
  \centering
  \small
  \setlength{\tabcolsep}{4.5pt}
  \begin{tabular}{lccc}
    \hline
    \textbf{System} & \textbf{BLEU}$\uparrow$ & \textbf{ROUGE-L}$\uparrow$ & \textbf{BERTSc.}$\uparrow$ \\
    \hline
    Ministral 0-shot & $12.37$          & $34.48$          & $46.36$          \\
    Ministral 8-shot & $16.44$          & $37.34$          & $50.69$          \\
    \panta-v2         & $\mathbf{20.30}$ & $\mathbf{41.19}$ & $\mathbf{55.00}$ \\
    \hline
    \panta-v2 (picto) & $14.61$          & $33.02$          & $47.42$          \\
    \hline
  \end{tabular}
  \caption{Surface metrics on the Instruct dataset. Questions were given as text, except for \panta-v2 (picto) where questions were given as pictograms.}
  \label{tab:metric-text-text}
\end{table}

\begin{table}[!h]
  \centering
  \small
  \setlength{\tabcolsep}{5pt}
  \begin{tabular}{lcccc}
    \hline
    \textbf{System} & \textbf{Rel.}$\uparrow$ & \textbf{Corr.}$\uparrow$ & \textbf{Simp.}$\uparrow$ & \textbf{Overall}$\uparrow$ \\
    \hline
    Ministral 0-shot & $\mathbf{0.861}$ & $\mathbf{0.913}$ & $0.732$          & $0.835$          \\
    Ministral 8-shot & $0.850$          & $0.900$          & $0.807$          & $0.852$          \\
    \panta-v2         & $0.846$          & $0.890$          & $\mathbf{0.823}$ & $\mathbf{0.853}$ \\
    \hline
    \panta-v2 (picto) & $0.685$          & $0.743$          & $0.789$          & $0.739$          \\
    \hline
  \end{tabular}
  \caption{LLM-judge scores: on the Dataset Instruct. Rel.: relevance; Corr.: correctness; Simp.: simplicity. Questions were given as text, except for \panta-v2 (picto) where questions were given as pictograms. }
  \label{tab:judge}
\end{table}

\begin{table}[!h]
  \centering
  \small
  \setlength{\tabcolsep}{8pt}
  \begin{tabular}{@{}lcc@{}}
    \hline
    \textbf{System} & \textbf{Correctness}$\uparrow$ & \textbf{Simplicity}$\uparrow$ \\
    \hline
    Ministral-3-8B (8-shot) & $\mathbf{0.850}$ & $0.788$          \\
    \panta-v2                & $0.806$          & $0.819$          \\
    \hline
    Reference answers       & $0.731$          & $\mathbf{0.919}$ \\
    \hline
  \end{tabular}
  \caption{Human evaluation of text answers.}
  \label{tab:human-text}
\end{table}

Table~\ref{tab:metric-text-text} shows that, on reference-based metrics, \textit{\panta-v2} outperforms Ministral-3-8B when the question is given as text.
This gain mainly reflects that the model has learned the style of the reference answers, which were produced by the same pipeline as its training data. 
When the question is given as pictograms, \textit{\panta-v2} remains close to the backbone answering text questions.

LLM judges give a more nuanced picture (Table~\ref{tab:judge}): Ministral-3-8B is at least as good as \textit{\panta-v2} on relevance and correctness, but not on simplicity. 
Human evaluation shows the same trend
(Table~\ref{tab:human-text}): \textit{\panta-v2} is slightly less correct than the 8-shot backbone (0.806 vs.\ 0.850) but simpler (0.819 vs.\ 0.788).
Reference answers are the simplest (0.919) but the least correct (0.731): our dataset construction favors simplicity over content, which likely explains why \textit{\panta-v2} loses some correctness compared to its backbone.
\textit{\panta-v2} sits between the backbone and its training references on both criteria, combining the backbone's knowledge with the training data's simplified style.

\subsubsection{Cross-Modal Human Evaluation}
\begin{table}[!h]
  \centering
  \small
  \setlength{\tabcolsep}{6pt}
  \begin{tabular}{@{}lcc@{}}
    \hline
    \textbf{Setting} & \textbf{Correctness}$\uparrow$ & \textbf{Transcription}$\uparrow$ \\
    \hline
    Text Q $\rightarrow$ Picto A & $0.53$ & $0.65$ \\
    Picto Q $\rightarrow$ Text A & $0.68$ & $0.59$ \\
    \hline
  \end{tabular}
    \caption{Human evaluation of \textit{\panta-v2} on cross-modal generation.}
  \label{tab:human-crossmodal}
\end{table}

Scores range from 0.53 to 0.68 (Table~\ref{tab:human-crossmodal}):
\textit{\panta-v2} handles pictogram inputs and outputs, with room for
improvement. In both settings, the loss occurs in the pictograms. With
pictogram questions, the text answers are correct (0.68), but the transcription
by RAPS loses information (0.59), as rare words and abstract concepts collapse
into generic pictograms. With text questions, the pictogram answers match the
text answers (0.65) but answer the question less well (0.53), as pictograms
convey less precise answers than text. Speech results are given in
Appendix~\ref{appendix:results-text-to-speech}.
\section{Conclusion}
\label{sec:conclusion}

We introduced FRAAC, a French dataset and multimodal model for pictogram-based augmentative and alternative communication. 
RAPS combines LLM-based concept extraction with retrieval over the ARASAAC vocabulary to align text with pictograms, yielding a synthetic corpus of 1.4M pairs, 300k of which are released. 
In a human ranking study, RAPS and \textit{\panta-v1}, trained on its data, are both ranked significantly above the rule-based state of the art.
These rankings also show that reference-based metrics separate systems of clearly different quality but give no reliable signal between close systems.
Finally, \textit{\panta-v2}, to our knowledge the first instruction-tuned multimodal model for French AAC users, answers text or pictogram questions in text, pictograms or speech. 
Evaluating the model with AAC users and speech therapists is the next step to assess its benefit for its target users.
\section{Limitations}
With the resources available, we could only rely on annotators from our
laboratory: none of our evaluations involve AAC users, so our results reflect
the judgments of these annotators, not the needs of people who rely on
pictogram-based communication. Involving AAC users and speech therapists will
be essential in future work.
Our experiments mainly rely on \texttt{gpt-oss-120b}, which requires H100-class GPUs and is costly in both compute and time.
More extensive ablations could improve both output quality and inference speed. 
For instance, we replaced the third stage of RAPS with direct LLM generation; it did not perform better, and we did not score this variant. 
 Our instruct dataset is mostly synthetic and was not systematically verified.
Finally, the model answers health-related questions for a vulnerable population that may not detect errors. It is a research prototype and must not be used as a source of medical advice without supervision.

\newpage
\bibliography{custom}
\appendix
\section{Appendix}
\subsection{Models-details}
\label{appendix:model-details}
\begin{table*}[!th]
  \centering
  \small
  \begin{tabular}{lp{5.5cm}p{3.5cm}r}
    \hline
    \textbf{Component} & \textbf{Description} & \textbf{Proposed Model} & \textbf{Parameters} \\
    \hline
    Pre-trained VLLM
      & Text-centric multimodal backbone based on a causal decoder architecture, responsible for cross-modal generation.
      & mistralai/Ministral-3-8B-Instruct-2512 \cite{ministral3}
      & 7{,}952.68M \\
    Pre-trained Image Encoder
      & Encodes visual inputs into a sequence of latent representations compatible with the language model.
      & Pixtral \cite{ministral3}
      & 403.31M \\
    Audio Codec
      & Converts continuous audio waveforms into discrete token sequences suitable for language model decoding.
      & BiCodec decoder part \cite{sparktts}
      & 92M \\
    Audio Decoder
      & Generates discrete audio tokens autoregressively.
      & Initialized with LLM weights
      & 1{,}359.04M \\
    Pictogram Decoder
      & Generates sequences of pictograms from latent representations using a causal decoding strategy.
      & Initialized with LLM weights
      & 928.53M \\
    \hline
  \end{tabular}
  \caption{\label{tab:components}
    Core components of the proposed multimodal architecture with parameter counts.}
\end{table*}


\definecolor{pictoFill}{HTML}{FBEEE8}
\definecolor{pictoLine}{HTML}{C2562F}
\definecolor{pictoText}{HTML}{6E2A12}
\definecolor{speechFill}{HTML}{EAF0F7}
\definecolor{speechLine}{HTML}{3E6EA3}
\definecolor{speechText}{HTML}{22415F}
\definecolor{coreFill}{HTML}{ECEAFA}
\definecolor{coreLine}{HTML}{6F67C9}
\definecolor{coreText}{HTML}{342D78}
\definecolor{neutFill}{HTML}{F2F1EC}
\definecolor{neutLine}{HTML}{8A8983}
\definecolor{neutText}{HTML}{2B2B29}

\begin{figure}[t]
  \centering
  \begin{tikzpicture}[
    x=1mm, y=1mm,
    font=\footnotesize,
    every node/.style={inner sep=0pt, outer sep=0pt},
    %
    modality/.style={
      draw=neutLine, line width=0.4pt, fill=neutFill, rounded corners=2pt,
      text=neutText, align=center, inner sep=1.6pt},
    picto/.style={modality, draw=pictoLine, fill=pictoFill, text=pictoText},
    speech/.style={modality, draw=speechLine, fill=speechFill, text=speechText},
    %
    frozen/.style={                      
      draw=coreLine!55, line width=0.35pt, fill=white, rounded corners=1.5pt,
      text=coreText, align=center, inner sep=1.5pt},
    trained/.style={                     
      draw=coreLine, line width=0.7pt, fill=coreFill, rounded corners=1.5pt,
      text=coreText, align=center, inner sep=1.5pt},
    vlmbox/.style={
      draw=coreLine, line width=0.6pt, fill=coreFill!35, rounded corners=3pt},
    detok/.style={
      draw=neutLine!80, line width=0.35pt, fill=neutFill, rounded corners=1.5pt,
      text=neutText, align=center, inner sep=1.2pt, font=\tiny},
    chip/.style={
      draw=pictoLine, line width=0.4pt, fill=white, rounded corners=1pt,
      minimum width=5.6mm, minimum height=5.6mm},
    pictocap/.style={font=\tiny, text=pictoText, align=center},
    flow/.style={-{Stealth[length=1.6mm, width=1.3mm]},
                 draw=neutLine, line width=0.45pt},
    inner/.style={-{Stealth[length=1.4mm, width=1.1mm]},
                  draw=coreLine!70, line width=0.4pt},
    side/.style={font=\scriptsize\itshape, text=neutText!85},
  ]

  \node[modality, minimum width=21mm, minimum height=13mm] at (-23.5,-6) (out-text) {};
  \node[text=neutText] at (-23.5,-2.8) {\textbf{Text}};
  \node[text=neutText, font=\scriptsize\itshape] at (-23.5,-8.4) {« À midi. »};

  \node[speech, minimum width=21mm, minimum height=13mm] at (0,-6) (out-speech) {};
  \node[text=speechText] at (0,-2.8) {\textbf{Speech}};
  \foreach \i/\h in {0/0.8, 1/1.7, 2/1.1, 3/2.4, 4/1.5, 5/2.6, 6/1.2, 7/1.9, 8/0.8} {
    \draw[speechLine, line width=0.5pt]
      ({-6.0+1.5*\i},{-8.6-\h}) -- ({-6.0+1.5*\i},{-8.6+\h});
  }

  \node[picto, minimum width=21mm, minimum height=13mm] at (23.5,-6) (out-picto) {};
  \node[text=pictoText] at (23.5,-2.8) {\textbf{Pictograms}};
  \node[chip] at (17.3,-8.8) (op1) {};
  \node[chip] at (23.5,-8.8) (op2) {};
  \node[chip] at (29.7,-8.8) (op3) {};
  \draw[pictoLine, line width=0.35pt] (op1.center) circle (1.4mm);
  \draw[pictoLine, line width=0.35pt] (op1.center) -- ([shift={(0,0.9)}]op1.center);
  \draw[pictoLine, line width=0.35pt] (op1.center) -- ([shift={(0.7,0)}]op1.center);
  \draw[pictoLine, line width=0.35pt] (op2.center) circle (0.8mm);
  \foreach \a in {0,60,...,300} {
    \draw[pictoLine, line width=0.3pt]
      ([shift={(\a:1.1mm)}]op2.center) -- ([shift={(\a:1.7mm)}]op2.center);
  }
  \draw[pictoLine, line width=0.35pt] (op3.center) circle (1.3mm);
  \draw[pictoLine, line width=0.35pt] (op3.center) circle (0.6mm);

  \node[detok, minimum width=21mm, minimum height=7.5mm] at (-23.5,-19) (dt-text)
    {text\\detokenizer};
  \node[detok, minimum width=21mm, minimum height=7.5mm] at (0,-19) (dt-speech)
    {audio codec\\decoder};
  \node[detok, minimum width=21mm, minimum height=7.5mm] at (23.5,-19) (dt-picto)
    {ARASAAC\\dictionary};

  \draw[flow] (dt-text.north)   -- (out-text.south);
  \draw[flow] (dt-speech.north) -- (out-speech.south);
  \draw[flow] (dt-picto.north)  -- (out-picto.south);

  \node[frozen, minimum width=21mm, minimum height=14mm] at (-23.5,-33) (d-text)
    {\textbf{LM head}\\[1.5pt]{\tiny of the LLM}};
  \node[trained, minimum width=21mm, minimum height=14mm] at (0,-33) (d-speech)
    {\textbf{Speech}\\[-1pt]\textbf{decoder}\\[1.5pt]{\tiny init.\ from LLM}};
  \node[trained, minimum width=21mm, minimum height=14mm] at (23.5,-33) (d-picto)
    {\textbf{Pictogram}\\[-1pt]\textbf{decoder}\\[1.5pt]{\tiny init.\ from LLM}};

  \draw[inner] (d-text.north)   -- (dt-text.south);
  \draw[inner] (d-speech.north) -- (dt-speech.south);
  \draw[inner] (d-picto.north)  -- (dt-picto.south);

  \node[vlmbox, minimum width=68mm, minimum height=29mm] at (0,-56.5) (vlm) {};
  \node[text=coreText, font=\scriptsize\itshape] at (0,-44.8)
    {pretrained vision--language model};

  \node[frozen, minimum width=60mm, minimum height=9mm] at (0,-52) (llm)
    {\textbf{LLM}\\[0.5pt]{\tiny shared multimodal representation}};

  \node[frozen, minimum width=28mm, minimum height=10mm] at (-16,-65) (emb)
    {text embeddings};
  \node[frozen, minimum width=28mm, minimum height=10mm] at ( 16,-65) (imgenc)
    {image encoder\\[0.5pt]{\tiny pictograms as images}};

  \draw[inner] (emb.north)    -- (-16,-56.5);
  \draw[inner] (imgenc.north) -- ( 16,-56.5);

  \draw[flow] (-23.5,-42) -- (d-text.south);
  \draw[flow] (  0  ,-42) -- (d-speech.south);
  \draw[flow] ( 23.5,-42) -- (d-picto.south);

  \node[modality, minimum width=32mm, minimum height=14mm] at (-18,-82) (in-text) {};
  \node[text=neutText] at (-18,-78.2) {\textbf{Text}};
  \node[text=neutText, font=\scriptsize\itshape] at (-18,-83.8) {« Je veux manger. »};

  \node[picto, minimum width=32mm, minimum height=14mm] at (18,-82) (in-picto) {};
  \node[text=pictoText] at (18,-78.2) {\textbf{Pictograms}};
  \node[chip] at ( 8.4,-83.2) (ip1) {};
  \node[chip] at (18.0,-83.2) (ip2) {};
  \node[chip] at (27.6,-83.2) (ip3) {};
  \draw[pictoLine, line width=0.35pt] ([shift={(0,1.35)}]ip1.center) circle (0.55mm);
  \draw[pictoLine, line width=0.35pt] ([shift={(0,0.8)}]ip1.center) -- ([shift={(0,-0.5)}]ip1.center);
  \draw[pictoLine, line width=0.35pt] ([shift={(-1.0,0.45)}]ip1.center) -- ([shift={(1.0,0.45)}]ip1.center);
  \draw[pictoLine, line width=0.35pt] ([shift={(0,-0.5)}]ip1.center) -- ([shift={(-0.8,-1.6)}]ip1.center);
  \draw[pictoLine, line width=0.35pt] ([shift={(0,-0.5)}]ip1.center) -- ([shift={(0.8,-1.6)}]ip1.center);
  \draw[pictoLine, line width=0.35pt, fill=pictoFill]
    ([shift={(0,-1.0)}]ip2.center)
      .. controls ([shift={(-1.4,0.4)}]ip2.center) and ([shift={(-0.6,1.3)}]ip2.center) .. ([shift={(0,0.3)}]ip2.center)
      .. controls ([shift={(0.6,1.3)}]ip2.center) and ([shift={(1.4,0.4)}]ip2.center) .. ([shift={(0,-1.0)}]ip2.center);
  \draw[pictoLine, line width=0.35pt] (ip3.center) circle (1.3mm);
  \draw[pictoLine, line width=0.35pt] (ip3.center) circle (0.6mm);

  \node[pictocap] at ( 8.4,-87.4) {je};
  \node[pictocap] at (18.0,-87.4) {vouloir};
  \node[pictocap] at (27.6,-87.4) {manger};

  \draw[flow] (in-text.north)  -- (-18,-71);
  \draw[flow] (in-picto.north) -- ( 18,-71);

  \end{tikzpicture}
  \caption{Overview of our French multimodal system. A pretrained
  vision language model encodes French text and pictogram sequences into a
  shared representation. Two transformer decoders, initialised from the
  \textsc{llm} and trained in parallel predict speech and pictogram tokens, while text is decoded by the \textsc{llm} itself.}
  \label{fig:overview}
\end{figure}
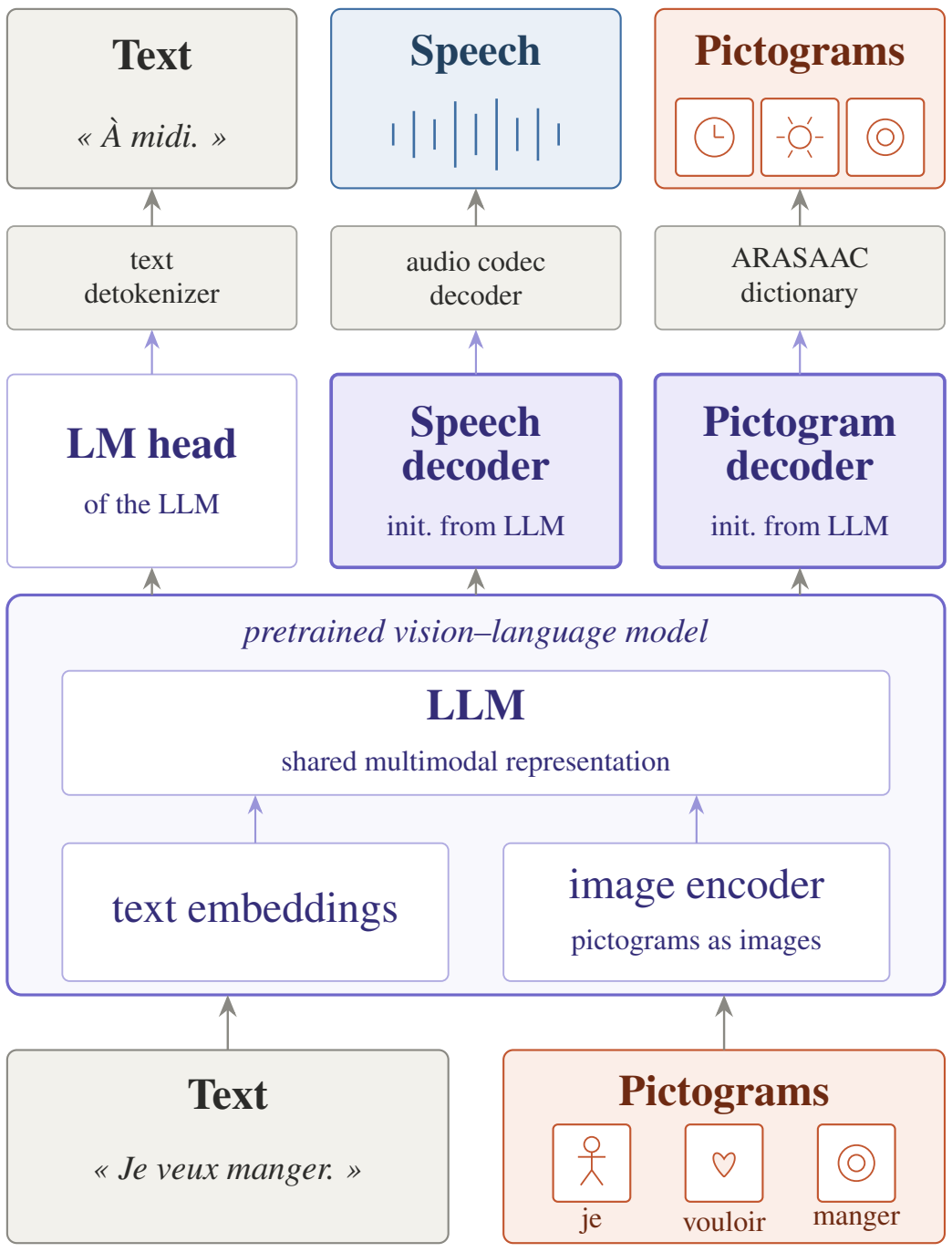
\vspace{-0.2cm}
The overall architecture is described in the Figure: \ref{fig:overview-panta}, and models parameters are describes in Table: 
\paragraph{Notation}
We consider a multimodal possible tokens $I = (I_1, \ldots, I_T)\in  (X, P, S) $ composed of  modalities: text tokens $X$, 
a sequence of pictographic representations $P$, and a speech representation $S$.
The objective of the model is, given a question represented in $X$ or $P$, to generate responses in all modalities I.
We denote by $d$ the hidden dimension of the LLM backbone, which all modalities are mapped into.
We denote by $H_\ell$ the hidden states of the LLM at layer $\ell$, by $\pi$ an instruction prompt.

\subsubsection{Modal Encoding}
\paragraph{Text.}
Text inputs are encoded using the standard token embedding layer of the LLM backbone.
\paragraph{Pictograms.}
Pictographic inputs are encoded using the image encoder of Pixtral, which is the vision 
encoder originally paired with our LLM backbone. Since Pixtral was designed to operate 
jointly with this backbone, its output dimension matches $d$ exactly, and no additional 
projection adapter is required.
In the original Pixtral setup, all patch representations are forwarded to the LLM, as the 
model is trained to handle complex natural images.
However, we argue that pictogram images are semantically simpler and that most patch 
representations are redundant.
We use Perceiver Resampler \cite{flamingo} the $N$ the number of patch embeddings produced by Pixtral into $K \ll N$ summary tokens where K is set to 4 in our setting.
Formally, let $\mathbf{P}_i \in \mathbb{R}^{N_p \times d}$ denote the $N_p$ patch tokens
produced by the vision encoder for the $i$-th pictogram, and let
$\mathbf{X} \in \mathbb{R}^{K \times d}$ be a set of $K$ learnable latent queries,
shared across all pictograms.

\begin{equation}
\begin{aligned}
    \mathbf{Q} &= \mathbf{X} \mathbf{W}_Q, \\
    \mathbf{K} &= \big[\mathbf{P}_i ; \mathbf{X}\big]\, \mathbf{W}_K, \\
    \mathbf{V} &= \big[\mathbf{P}_i ; \mathbf{X}\big]\, \mathbf{W}_V, \\
    \tilde{\mathbf{X}} &= \mathbf{X}
        + \operatorname{softmax}\!\Big( \tfrac{\mathbf{Q}\mathbf{K}^{\top}}{\sqrt{d}} \Big) \mathbf{V}, \\
    \mathbf{X} &= \tilde{\mathbf{X}}
        + \operatorname{FFN}\big(\tilde{\mathbf{X}}\big),
\end{aligned}
\label{eq:perceiver_resampler}
\end{equation}
where $[\,\cdot\,;\,\cdot\,]$ denotes concatenation along the token dimension,
$\mathbf{W}_Q, \mathbf{W}_K, \mathbf{W}_V \in \mathbb{R}^{d \times d}$
are learned projection matrices, and $\operatorname{FFN}$ is a two-layer MLP.
The final representation of the $i$-th pictogram is
$\mathbf{Z}_i = \mathbf{X} \in \mathbb{R}^{M \times d}$,
i.e.\ a fixed number $M$ of tokens independent of $N_p$.
For clarity, we omit multi-head attention and layer normalization from the notation.
This operation produces $K$ pooled tokens per pictogram, which are then concatenated and fed to the LLM backbone.

\subsubsection{Modality-Specific}
\label{par:modality_decoding}
Modality-specific generation is handled by lightweight shallow decoders operating in an early-exit fashion: at a selected intermediate layer $\ell$ of the backbone, the hidden 
states $H_\ell$ are branched off and fed into dedicated decoders, one per output modality, while the main backbone continues processing for text generation.

Each modal decoder operates autoregressively, where the task is reduced to token prediction.

\paragraph{Pictogram generation.}
The pictogram decoder predicts, at each step, the identifier of a pictogram in the ARASAAC\footnote{\url{https://arasaac.org/about-us}} vocabulary.
ARASAAC is an open-access repository of graphical communication resources designed to support AAC.
The project is funded by the Department of Education, Culture and Sports of the Government of Aragon and distributes its resources under a Creative Commons BY-NC-SA license.

\paragraph{Speech generation.}
For speech generation, we follow the approach described in \cite{sparktts}: the decoder was trained to predict the discrete semantic tokens of the BiCodec tokenizer vocabulary \cite{sparktts}, which can be converted to a waveform by the BiCodec decoder.

\subsection{Implementation details}
\label{appendix:implementation-details}
\subsubsection{Training Overview}
\paragraph{Stage 1: modality alignment.}
Following standard multimodal training practice~\cite{mini-omni}, the
first stage aligns the representations of the new modalities with those
of the \textit{Ministral-3-8B-Instruct} language model part. The LLM is kept
frozen, so that the model acquires the new modalities without losing its
text instruction-following abilities. Each new component is trained
independently on its own task: the pictogram decoder on text-to-pictogram
transcription, the speech decoder on text-to-speech, and the pictogram
encoder on pictogram-to-text. The pictogram encoder is built on the
Pixtral vision encoder only the adapter layer connecting Pixtral and the Perceiver Resampler are trained.

\paragraph{Stage 2: multitask instruction tuning.} The second stage combines
the modality alignment tasks with our Instruct Dataset
(Section~\ref{sec:instruct_dataset}), for a total of five tasks:
text-to-speech, text-to-pictogram, pictogram-to-text, and question answering
from a text or a pictogram question, with answers in text, speech and
pictograms. At each step, one task is sampled uniformly, and only the parameters involved in that task are updated. The pictogram
encoder remains frozen throughout this stage, and the LLM backbone is updated
only when the target includes text.

\subsubsection{Training Objective}
All tasks for the training loss function can be reduced to token prediction in an auto-regressive manner, conditioned on different token-level. This is why we rely on the classical Cross-Entropy (CE) loss.

In the first stage for tasks involving a non-text target modality (e.g., text-to-pictogram or text-to-speech translation), the prompt $\pi$ instructs the model to translate the input into the target modality.
For tasks with a text target (pictogram-to-text translation), the prompt instructs the model to produce a textual translation hence this fixed prompt strategy should not impact the core LLM ability to deal with a vast variety of prompts. It is important to notice that in this paper we not use a global prompt system.
In the second stage, the prompt instructs the model to answer the question, given as text or as a pictogram sequence, with a single simple sentence.

\paragraph{Pictogram generation loss.}
$H_\ell$ denotes the hidden state extracted at layer $\ell = -4$ (i.e., the fourth-to-last layer). The model is conditioned on the LLM-generated hidden space which is itself conditioned by the prompt $\pi$ and input sequence $I$.

\begin{equation}
     \mathcal{L}_P = -\sum_{t=1}^{T} \log P\!\left(O_t \mid \, H_\ell,\, O_{<t}\right)
     \label{eq:loss_picto}
\end{equation}

\paragraph{Speech generation loss.}
Let $\mathbf{s}$ denote the speaker embedding computed by the BiCodec encoder, and $H_\ell$ denotes the hidden state extracted at layer $\ell = -6$. The model is conditioned on the LLM-generated hidden space which is itself conditioned by the prompt $\pi$ 
 and the input representation $I$. The speech generation loss is:
 \begin{equation}
     \mathcal{L}_S = -\sum_{t=1}^{T} \log P\!\left(O_t \mid H_\ell,\, O_{<t},\, \mathbf{s}\right)
     \label{eq:loss_speech}
 \end{equation}

 \paragraph{Text generation loss.}
 \begin{equation}
     \mathcal{L}_X = -\sum_{t=1}^{T} \log P\!\left(O_t \mid \pi,\, O_{<t},\, I\right)
     \label{eq:loss_text}
 \end{equation}

\paragraph{Instruct Loss}
The loss from the instruct model is the mean of all previous losses

\begin{equation}
    \mathcal{L}_{Instruct} = \frac{\mathcal{L}_X + \mathcal{L}_S + \mathcal{L}_P}{3}
\end{equation}

\subsubsection{Implementation Details with details}
\begin{figure*}[!t]
\centering
\includegraphics[width=\textwidth]{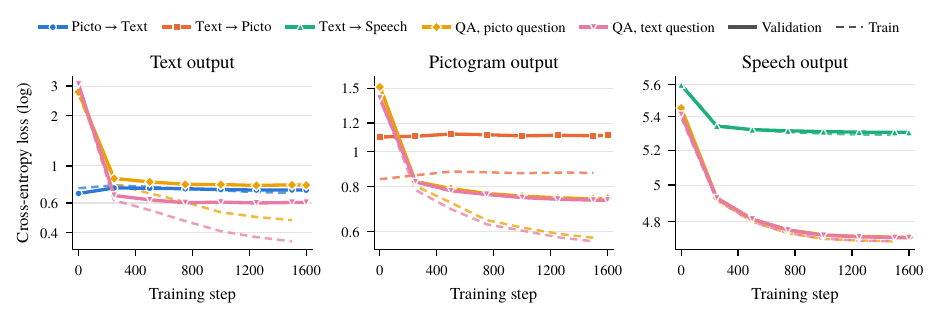}
\caption{Stage 2 training curves of the selected run (LLM learning rate
$10^{-6}$, decoders $5\cdot10^{-6}$), grouped by output modality. Solid lines:
validation loss; dashed lines: training loss on an evaluated subset. The
y-axis is in log scale.}
\label{fig:instruct_curves}
\end{figure*}

\begin{table*}[!ht]
\centering
\small
\begin{tabular}{@{}lccc@{}}
\toprule
 & \multicolumn{2}{c}{\textbf{Stage 1}} & \textbf{Stage 2} \\
\cmidrule(lr){2-3}\cmidrule(l){4-4}
 & Picto$\rightarrow$Text & Text$\rightarrow$Picto & Instruct \\
\midrule
Initialization    & Ministral-3-8B-Instruct & Ministral-3-8B-Instruct & \panta-v1 \\
Trained modules   & Adapter + Perceiver     & Pictogram decoder       & LLM + decoders \\
GPUs (H100)       & 4                       & 4                       & 8 (FSDP) \\
Global batch size & 256                     & 256                     & 64 \\
Steps             & 24k                     & 20k                     & 1.6k \\
Warmup steps      & 2k                      & 2k                      & 80 \\
$\beta_2$         & 0.999                   & 0.999                   & 0.95 \\
Min. LR           & 5e-7                    & 5e-7                    & 1e-7 \\
\midrule
\multicolumn{4}{@{}l}{\textit{Learning-rate grid (selected value in bold)}} \\
Picto$\rightarrow$Text  & \multicolumn{3}{l}{\textbf{5e-5}, 1e-4, 2e-4, 5e-4, 1e-3, 2e-3, 5e-3} \\
Text$\rightarrow$Picto  & \multicolumn{3}{l}{2e-5, 5e-5, \textbf{1e-4}, 5e-4} \\
Instruct, LLM           & \multicolumn{3}{l}{7e-7, \textbf{1e-6}, 2e-6, 1e-5, 2e-5, 5e-5, 1e-4} \\
Instruct, decoders      & \multicolumn{3}{l}{2e-6, \textbf{5e-6}, 1e-5, 2e-5, 5e-5, 1e-4} \\
\bottomrule
\end{tabular}
\caption{Training hyperparameters. Stage 2 trains on five
tasks (picto-to-text, text-to-picto, text-to-speech, and question answering
from a text or a pictogram question), sampled uniformly with oversampling of
the smaller datasets; the LLM is frozen for text-to-picto and text-to-speech
batches.}
\label{tab:hyperparams}
\end{table*}

All components of the model are trained independently with the cross-entropy loss, following the teacher-forcing paradigm, since the shared backbone remains frozen throughout training.
Throughout the first stage, the backbone LLM remains frozen, which allows all other components to be trained under different configurations while preserving the instruction-following abilities of the original model.
At the beginning of training, the \textit{Ministral-3-8B-Instruct} model, originally released in FP8, is cast to bfloat16.
The same conversion is applied to the other pre-trained components, with the difference that, unlike the backbone, these components are subsequently trained.
For the first stage we used the AdamW optimizer~\cite{adamw} with $\beta_1 = 0.9$ and $\beta_2 = 0.999$ (except for text-to-speech, where $\beta_2 = 0.96$), a cosine learning rate scheduler with a 10\% warm up ratio, a weight decay of 0.01, and gradient clipping set to 1.

For the second stage, we train for 1,600 steps, which corresponds to about one
pass over the Instruct Dataset for each question-answering task, using AdamW
\cite{adamw} with $\beta_1 = 0.9$ and $\beta_2 = 0.95$ for all tasks.
Table~\ref{tab:hyperparams} summarizes the hyperparameters of both stages.
Learning rates are selected by grid search, and checkpoints are chosen on the
validation loss; a limitation of this criterion is that a lower loss does not
guarantee better generations.
Finally we select \textit{\panta-v2} with the last checkpoint with the learning rate of $1e-6$ for the backbone and $5e-6$ the training curves are reported in Figure: \ref{fig:instruct_curves}

\subsection{Data Preprocessing RAPS}
\label{appendix:data-preprocessing-raps}
Once collected, the raw text data was processed through a standardized pipeline designed to ensure linguistic quality and stylistic consistency across sources:
\begin{enumerate}
    \item \textbf{Translation}: Non-French content was automatically translated into French using the neural machine translation model \textit{facebook/nllb-200-distilled-600M}~\cite{nllb}.
    \item \textbf{Sentence segmentation}: The text was split into individual sentences using \textit{Stanza}~\cite{stanza}, a robust multilingual NLP toolkit.
    \item \textbf{Rule-based cleaning}: Each sentence was cleaned using regular expressions to remove noise such as special characters, malformed punctuation, or formatting artifacts inherited from web scraping.
    \item \textbf{Semantic filtering}: \textit{Qwen3-14B} ~ \cite{qwen3} was used as a judge quote to assess each sentence and discard those that were too complex, ambiguous, or otherwise unsuitable for pictogram-based representation.
\end{enumerate}
 \subsection{Dataset Instruct}
\label{appendix:instruct_dataset}
\paragraph{Synthetic subset.}
Generation is driven by a taxonomy of 4 themes, 29 sub-themes and 3 age-based personas, 
with 95\% direct answers and 5\% referrals to a relevant interlocutor (doctor, antaarent, teacher). 
Pairs are filtered with rule-based checks, semantic deduplication and a blind LLM-as-a-judge.

\paragraph{\anonymousdataset and French-Instruct subsets.}
Questions are extracted from \anonymousdataset and French-Instruct, filtered by rules and an LLM judge 
(self-contained, general knowledge, no named entities), then answered in a single sentence of 
4 to 16 words. For French-Instruct, the original answer of the corpus is given to the model as 
a reference to rephrase. A final filter removes inconsistent question/answer pairs.

\paragraph{Multimodal conversion.}
All subsets are first built as text-to-text question answering pairs. Each text answer is then converted into a pictogram sequence and into speech using automatic text-to-pictogram and text-to-speech approaches, respectively. 

\subsection{Ground truth datasets}
\label{appendix:ground-true-dataset}
\begin{enumerate}
    \item \textbf{Medical}~\cite{norre2025thesis,bouillon2017babeldr}: 250 simple
          doctor-to-patient questions from BabelDr, translated into ARASAAC
          by a speech therapist and by the researchers.
    \item \textbf{Email}~\cite{norre2021extending,sevens2017simplified}: 130 sentences from
          WAI-NOT, a platform used by people with an intellectual
          disability, their teachers and parents. Originally Dutch, they
          were translated into French and ARASAAC by the researchers.
          They are longer and more naturalistic than the Medical ones.
    \item \textbf{Book}~\cite{vaschalde2018automatic,norre2021extending}: 254
          sentences from six copyright-free children's stories, the longest
          and most varied.
    \item \textbf{Arasaac materials}: 566 sentences from 33 educational materials (stories, songs, activity sheets) published by teachers and speech therapists in the ARASAAC catalogue\footnote{\url{https://arasaac.org/materials}, CC BY-NC-SA 4.0.}. The pictogram sequences were written by the material authors themselves. We extracted them from the PDFs and resolved each pictogram to its ARASAAC id (CLIP retrieval, with vision-language-model re-annotation for rasterised pages and manual checking). The source texts are mostly Spanish (20 materials) or English (10), so the French sentences were machine-translated from the source language with \textit{Claude Opus~5}.
\end{enumerate}

\begin{table}[!h]
  \centering
  \small
  \begin{tabular}{lr}
    \hline
    \textbf{Subset} & \textbf{\# Examples} \\
    \hline
    Train (Synthetic)         & $14\,965$ \\
    Train (\anonymousdataset) & $4\,529$  \\
    \hline
    \textit{Total train}      & $19\,494$ \\
    Test (French-Instruct)    & $1\,285$  \\
    \hline
  \end{tabular}
  \caption{\label{tab:Panta-instruct}Composition of the Instruct dataset.}
\end{table}

\subsection{Text to Speech}

\subsubsection{Dataset Text-Speech}
\begin{table*}[!th]
\centering
\small
\begin{tabular}{lrrr}
\hline
\textbf{Dataset} & \textbf{Total (kh)} & \textbf{Filtered (kh)} & \textbf{\#Segments} \\
\hline
MLS~\cite{MLS}             & 0.99  & 0.83  & 216\,563 \\
EPAC~\cite{epac}           & 1.46  & 1.01  & 375\,936 \\
Audiocite~\cite{audiocite} & 4.76  & 4.35  & 573\,808 \\
Yodas~\cite{yodas}         & 15.29 & 6.46  & 1\,082\,094 \\
\hline
\textbf{Total}             & 22.50 & 12.65 & 2\,248\,401 \\
\hline
\end{tabular}
\caption{Statistics of the text-to-speech training data, before and after filtering. Durations are given in thousands of hours (kh).}
\label{tab:data-text-to-speech}
\end{table*}

The purpose of this dataset is to provide cleaned text/speech pairs in which each example is
spoken by a single voice. This constraint follows from the speech generation task: the model
is trained to produce speech from a textual, and therefore has to predict not only the linguistic content but also the acoustic identity that carries it. When the speaker changes within a training example, that identity is no longer predictable from the input, and the model can only average over the variation it observes, which degrades the
consistency of the generated voice. 
To obtain aligned speech/text data, we follow a processing pipeline inspired by WhisperX \cite{whisperx}.
The pipeline proceeds as follows:
\begin{enumerate}
    \item \textbf{Voice Activity Detection:} Raw audio recordings are segmented into short chunks using \textit{Silero VAD} \cite{Silero}.
    \item \textbf{Speaker Diarization:} Each chunk is further processed with \textit{pyannote} \cite{Pyannote} to ensure that every segment contains a single speaker.
    \item \textbf{Duration Splitting:} Segments longer than 30\,s, the input limit of our Whisper encoder, are split at the lowest-energy point between 25\,s and 30\,s, so that the cut falls in a silence rather than mid-utterance. The operation is repeated until all segments are under 30\,s.
    \item \textbf{Transcription:} Transcriptions are generated using both \textit{openai/whisper-large-v3-turbo} \cite{Whisper} and the Canary ASR model \cite{canary}.
    \item \textbf{Consistency Filtering:} As no ground-truth transcription is available, we transcribe each segment with
    two independent systems. Canary \cite{canary} was selected for its favourable
    accuracy/speed trade-off, being among the top-ranked systems on the Open ASR
    Leaderboard\footnote{\url{https://huggingface.co/spaces/hf-audio/open_asr_leaderboard},
    accessed \today.} while achieving the highest inference throughput. Whisper
    \cite{Whisper} was selected as it has already been used to transcribe
    hundreds of hours of French audio \cite{pelloin2026data}.
   Their disagreement is used as a quality proxy: taking the Canary output as reference and the Whisper output as hypothesis, we retain only the segments whose World Error Rate (WER) is below 10\%. Since the corpus is intended for speech synthesis, which requires clean audio, we
    assume that a disagreement between two strong ASR systems indicates a degraded or noisy signal, and we use this criterion to discard such segments.We set the threshold at 10\%, by reference to the WER typically reported by state-of-the-art systems on French speech \cite{pantagruel}. Since this value corresponds to the error rate of a single system against a human reference, two independent systems would be expected to disagree by at least as much on average;
    requiring their mutual WER to stay below 10\% therefore retains only segments on which both systems are markedly more consistent than usual, which we take as an indication of good signal quality.
    \item \textbf{Music Detection:} Segments containing music are identified using he music detector: \textit{ina-foss/ssl-music-detection-music2vec} \cite{pantagruel}.
\end{enumerate}

For \textit{text-to-speech}, higher speech quality is required; therefore, segments flagged by the music detector are discarded.
We apply this pipeline to the MLS \cite{MLS}, EPAC \cite{epac}, YODAS
\cite{yodas}, and Audiocite \cite{audiocite} corpora. These sources were chosen
for their complementary in speakers, speech types, and recording conditions:
MLS and Audiocite provide a large volume of clean read speech, EPAC adds
broadcast speech from radio archives, and YODAS contributes in-the-wild YouTube
recordings of varying quality. Statistics for the resulting \textit{text-to-speech}
dataset are reported in Tables~\ref{tab:data-text-to-speech}.

\subsubsection{Speech evaluation metrics}
\label{appendinx:speech-evaluation-metrics}
We used three metrics for speech evaluation MOS, Speaker Similarity metric, and Word Error Rate WER.
\begin{enumerate}
\item \textbf{MOS}: a neural-network-based metric that predicts the Mean Opinion Score
(MOS) of an audio signal without requiring a reference, aiming to approximate human
subjective quality ratings (range: typically $1$ to $5$, higher is better). We use UTMOS-v2 \cite{utmosv2}
\item \textbf{Speaker Similarity}: cosine similarity between embeddings obtained from the
\texttt{speechbrain/spkrec-ecapa-voxceleb} model \cite{Ecapa}.
\item \textbf{WER}: measures the intelligibility of the generated speech. Each audio output is
transcribed with an automatic speech recognition (ASR) model, here Whisper large-v3~\cite{Whisper},
and the transcription is compared to the target text. The WER is computed as
\begin{equation}
\mathrm{WER} = \frac{S + D + I}{N},
\end{equation}
where $S$, $D$ and $I$ are the numbers of word substitutions, deletions and insertions, and $N$
is the number of words in the reference text (lower is better). Before scoring, both the reference
and the transcription are normalized.
\end{enumerate}

We transcribe the generated speech with
Whisper-large-v3 \cite{Whisper} and compute the WER against the text the speech
is meant to convey. For Voxtral \cite{voxtral}, a text-to-speech model, this is
the input text it reads, taken from our Instruct Dataset
(Section~\ref{sec:instruct_dataset}). For \textit{\panta-v2}, it is the text
answer the model generates for the same question.

\subsubsection{Audio tokenizer}
\begin{table}[t]
\centering
\resizebox{\columnwidth}{!}{%
\begin{tabular}{lcccccc}
\hline
\textbf{Tokenizer} &
\textbf{PESQ}$\uparrow$ &
\textbf{STOI}$\uparrow$ &
\textbf{Mel}$\downarrow$ &
\textbf{STFT}$\downarrow$ &
\textbf{MOS}$\uparrow$ &
\textbf{Spk.}$\uparrow$ \\
\hline
Ground Truth & 4.644 & 1.000 & 0.000 & 0.000 & 2.439 & 1.000 \\
\hline
BiCodec      & 2.086 & 0.869 & 5.532 & 0.214 & 2.604 & \textbf{0.771} \\
BiCodec FT   & \textbf{2.213} & \textbf{0.888} & \textbf{5.273} & \textbf{0.202} & \textbf{2.681} & 0.733 \\
\hline
\end{tabular}%
}
\caption{Reconstruction quality of the BiCodec tokenizer, before and after fine-tuning (FT). Spk.: speaker similarity. Best codec results in bold.}
\label{tab:tokenizers}
\end{table}

One of the main drawbacks of the BiCodec \cite{sparktts} is that the model is trained on English and Chinese data only, whereas the target language is French.
We continued the training on the MLS~\cite{MLS} and EPAC~\cite{epac} datasets, following the loss defined in the original paper \cite{sparktts}.
To evaluate our model, we used a subsample of 1000 wav files extracted from Yodas~\cite{yodas} to evaluate our new tokenizer using the following metrics:
\subsection{Audio tokenizer metrics}
\label{appendix:audio}
\begin{enumerate}
\item \textbf{Mel Distance (Reconstruction)}: L1 distance between the mel-scaled
magnitude spectrograms of the ground truth and the generated sample.
\item \textbf{STFT Distance (Reconstruction)}: L1 distance between time-frequency
representations of the ground truth and the prediction, computed using a multi-scale
Short-Time Fourier Transform (STFT).
\item \textbf{PESQ (Perceptual Evaluation of Speech Quality)}: an objective metric that
estimates the perceived quality of a speech signal by comparing it to a clean reference.
It uses a psychoacoustic model to approximate how humans perceive distortions (range:
$-0.5$ to $4.5$, higher is better).
\item \textbf{STOI (Short-Time Objective Intelligibility)}: an objective intelligibility
metric that measures how understandable a degraded speech signal is compared to a clean
reference, based on short-time envelope correlations across frequency bands (range: $0$
to $1$, higher is better).
\end{enumerate}

All results are reported in Table~\ref{tab:tokenizers}.
In a subsequent stage, we train only the decoder of the tokenizer on the single-speaker dataset, again with the goal of producing a single consistent output voice.

\subsubsection{Training Text-to-Speech}
\paragraph{First Stage: Text-to-Speech}
The text-to-speech components are trained in two stages to predict the semantic tokens of our audio tokenizer, using the loss defined in Equation~\ref{eq:loss_speech} and the dataset described in Table~\ref{tab:data-text-to-speech}.
In the first stage, transformers decoder layers are trained, using a learning rate of $5e^{-4}$.
In the second stage, we fine-tune the model on the Synpaflex corpus~\cite{synpaflex}, a French single-speaker reading open dataset, in order to mitigate the potential misuse of the model for generating synthetic speech that could be used for voice impersonation.

\paragraph{Audio Tokenizer Training.}
Our audio tokenizer, BiCodec~\cite{sparktts}, is originally trained on English and Chinese data.
We perform continued pre-training on the MLS~\cite{MLS} and EPAC~\cite{epac} datasets, following the loss defined in the original paper.

\paragraph{Results text-to-speech}
\label{appendix:results-text-to-speech}
For speech (Table~\ref{tab:speech}), \textit{\panta-v2} matches \textit{Voxtral} \cite{voxtral} in intelligibility and naturalness.
At the same time, its much higher speaker similarity shows that the voice is successfully adapted to the target speaker. 
These scores are unaffected by the input modality. .
\begin{table}[!h]
  \centering
  \small
  \setlength{\tabcolsep}{5pt}
  \begin{tabular}{llccc}
    \hline
    \textbf{System} & \textbf{Input} & \textbf{Spk. sim.}$\uparrow$ & \textbf{WER}$\downarrow$ & \textbf{MOS}$\uparrow$ \\
    \hline
    Voxtral  & Text  & $0.168$          & $\mathbf{1.89}$ & $\mathbf{3.32}$ \\
    \panta-v2 & Text  & $\mathbf{0.582}$ & $2.27$          & $3.17$          \\
    \panta-v2 & Picto & $0.581$          & $2.41$          & $3.17$          \\
    \hline
  \end{tabular}
  \caption{Evaluation of text-to-speech.}
  \label{tab:speech}
\end{table}

\clearpage
\onecolumn
\raggedbottom   

\subsection{Prompts}
\label{appendix:prompts}

This section lists the prompts used in our pipelines. Text in
\ph{angle brackets} is a placeholder filled at run time; JSON schemas
constrain the output format through structured generation.

\subsubsection{Retrieval-Augmented Pictogram Selection (RAPS)}
\label{app:prompts-raps}

RAPS uses four prompts, all run with \texttt{gpt-oss-120b}.
Prompt~\ref{prompt:raps-concepts} extracts the
concept candidates that drive retrieval (Stage~1).
Prompts~\ref{prompt:raps-replace}, \ref{prompt:raps-add}
and~\ref{prompt:raps-remove} edit the rule-based draft sequence one
operation at a time (Stage~3), each operation receiving the output of the
previous one. In Stage~3, pictogram candidates are given as JSON entries
with their \texttt{lemma}, \texttt{picto\_id}, \texttt{definition} and
visual \texttt{description}.

\begin{promptbox}{RAPS Stage~1 --- Concept extraction (\texttt{gpt-oss-120b})}{prompt:raps-concepts}
\begin{Verbatim}
Objective: convert the text to a sequence of pictograms for Augmentative and Alternative Communication (AAC).

- Create a list of concepts (words or groups of words) that would be useful for representing the meaning of the text
- Use frequent and simple words likely to appear in a reduced lexicon
- Add synonyms and split versions of complex concepts
- Add hypernyms or more generic concepts for named entities
- Add salient properties for named entities or concepts that appear multiple times in the text
- Add split versions of numbers
- Add variants with negation and antonyms when appropriate
- Always represent tense as a time marker (past, future, ...) before the verb in base form. Do not add a marker for present tense. Never use conjugated auxiliaries or inflected verbs. Example: 'j'ai mangé' → 'past eat', 'je mange' → 'eat'
- If the text is a question, always add an explicit 'question' marker. Example: 'Tu manges ?' → 'you eat question'
- Add explicit modality, negation, location, manner markers if needed
- Replace expressions with their idiomatic meaning
- For a polysemous word, pick the lemma and definition for the sense meant in this sentence, not other senses of the same word
- When the text names a bodily sensation, symptom or medical state (pain, heat, cold, swelling, itching, fatigue, ...), write the definition so it explicitly states a sensation felt by a person's body or a medical condition, never a generic physical, weather or object sense of the same word: a pictogram lexicon typically has several entries sharing one English word, one for each register, and only the definition tells the retrieval which one is meant.
- When a French expression names a single medical symptom or idiom, generate it as one symptom concept with synonym variants, not as two unrelated halves: do not additionally add the generic body part on its own when the symptom concept already implies it.
- A precise medical or technical term (an uncommon treatment, diagnostic procedure, or specialized condition) may have no exact entry in a small pictogram lexicon. When that looks likely, also generate its broader medical category and the relevant specialist or equipment, so retrieval still has a relevant candidate instead of falling back to something that only shares a word root or substring with the term but means something unrelated.
- Generate a sequence of English lemmas with corresponding definitions for all concept candidates.
- Do not generate more than 5 alternatives for each input content word

Examples:
- John => person, individual, masculine (do not use first_name if referencing a person)
- Mickael Jackson => person, singer, music, USA
- Paris => city, capital, France
- 12345 => 12000, 12, thousand, 300, 45
- detach => not attach
- j'ai mangé => past, eat
- Tu manges ? => you, eat, question
- Comment vas-tu ? => you, be (feel), how, question (not go: idiomatic sense, not travel)
- il a des sueurs nocturnes => night sweat (a bodily sensation of sweating during sleep, not outdoor/weather heat), sweat (the definitions must describe a body sensation, not thermal energy, weather or a liquid)
- elle a des douleurs articulaires => joint pain, ache (one pain symptom; do not also add a separate plain "joint" anatomy concept, it is already implied)
- ses pieds gonflent à cause de la chaleur => swell, oedema (a bodily accumulation of fluid under the skin — not "inflate/blow up", which describes forcing air into an object)

Output a single JSON list with the schema:
\ph|JSON schema: lemma + definition^

Input text (in \ph|source language^):
\ph|source sentence^
\end{Verbatim}
\end{promptbox}

\begin{promptbox}{RAPS Stage~3a --- Sequence composition: \textsc{Replace} (\texttt{gpt-oss-120b})}{prompt:raps-replace}
\begin{Verbatim}
You are correcting a first-draft pictogram sequence for Augmentative and Alternative Communication (AAC), one operation at a time. This is step 1 of 3: REPLACE.
GENERAL RULES:
1. The source text is always correct as written. Never reinterpret, correct, or paraphrase it before analysing the sequence. If a word seems unusual, treat it as intentional and work from its literal meaning.

REPLACE RULES
1. Determine what the entry is meant to render in the source sentence. The `lemma` gives a first orientation, but the sentence is decisive: the same lemma can surface in several senses, and the surrounding context pins down which one. Where no candidate matches the lemma exactly, do not discard candidates on that basis alone — one whose `definition` fits the intended sense is a valid alternative.
2. From the candidate list, select the three entries that could plausibly render that fragment. Judge by `definition`, not by `lemma`: a superficially unrelated lemma may carry precisely the right sense, while a matching lemma may not. The proposed entry is always among the three you compare.
3. Choose between them on three grounds, in this order:
   - SEMANTIC FIT — does the `definition` match the sense the sentence uses? A definition that names a different sense disqualifies the entry, however well it is drawn.
   - DEPICTION — among entries whose definitions fit, does the `description` show something a reader would recognise quickly? Prefer a concrete, recognisable scene over an abstract symbol, and a depiction that could only mean this over one that could mean several things.
   - CONTEXT — does the depiction fit the situation the rest of the sentence describes?
   Do not reason on `lemma` at this stage — it plays no role in choosing between candidates.

4. Take whichever entry wins this comparison. If that is the proposed one, keep it unchanged. Only make a replacement when the alternative is a clearly better or more natural representation of the intended meaning.

Never output two pictograms for the same lemma.

EXAMPLES

Wrong sense revealed by the depiction:
Text: "j'ai mal au ventre depuis ce matin"
Proposed sequence:
[{"lemma": "I", "picto_id": "200"}, {"lemma": "have", "picto_id": "205", "description": "An open hand holding a small object in its palm."}, {"lemma": "pain", "picto_id": "201", "description": "A grinning demon face with horns, used to depict evil or wickedness."}, {"lemma": "belly", "picto_id": "202", "description": "A human torso with the stomach area highlighted in red."}, {"lemma": "morning", "picto_id": "203"}]
Candidates:
[{"lemma": "I", "picto_id": "200"}, {"lemma": "have", "picto_id": "205", "description": "An open hand holding a small object in its palm."}, {"lemma": "pain", "picto_id": "201", "description": "A grinning demon face with horns, used to depict evil or wickedness."}, {"lemma": "pain", "picto_id": "204", "description": "A person with a pained expression holding their side, with red radiating lines at the sore spot."}, {"lemma": "belly", "picto_id": "202", "description": "A human torso with the stomach area highlighted in red."}, {"lemma": "morning", "picto_id": "203"}]
Replaced:
[{"lemma": "I", "picto_id": "200"}, {"lemma": "have", "picto_id": "205"}, {"lemma": "pain", "picto_id": "204"}, {"lemma": "belly", "picto_id": "202"}, {"lemma": "morning", "picto_id": "203"}]
(201 and 204 share the lemma "pain", but a demon face depicts moral evil, an unrelated sense, so 201 is replaced by 204; "have" is left unchanged, removing it is a later step's job)

Same lemma, clearer depiction:
Text: "il fait très chaud aujourd'hui"
Proposed sequence:
[{"lemma": "make", "picto_id": "182"}, {"lemma": "hot", "picto_id": "180", "description": "A thermometer with red mercury rising to the top, indicating high temperature."}]
Candidates:
[{"lemma": "make", "picto_id": "182"}, {"lemma": "hot", "picto_id": "181", "description": "A yellow sun with wavy red heat lines radiating outward."}, {"lemma": "hot", "picto_id": "180", "description": "A thermometer with red mercury rising to the top, indicating high temperature."}]
Replaced:
[{"lemma": "make", "picto_id": "182"}, {"lemma": "hot", "picto_id": "181"}]
(the sun with heat lines is immediately recognisable, a thermometer is a conventional instrument the reader has to interpret; "make" is left unchanged)

Output a single JSON list, with the proposed sequencz in the same order, using the schema:
\ph|JSON schema: lemma + picto\_id^

Original text (in \ph|source language^):
\ph|source sentence^

Proposed sequence:
\ph|rule-based draft sequence^

Candidates:
\ph|retrieved candidates^
\end{Verbatim}
\end{promptbox}

\begin{promptbox}{RAPS Stage~3b --- Sequence composition: \textsc{Add} (\texttt{gpt-oss-120b})}{prompt:raps-add}
\begin{Verbatim}
You are correcting a pictogram sequence for Augmentative and Alternative Communication (AAC), one operation at a time. This is step 2 of 3: ADD.

GENERAL RULES:
1. The source text is always correct as written. Never reinterpret, correct, or paraphrase it before analysing the sequence. If a word seems unusual, treat it as intentional and work from its literal meaning.

Add what is missing. Never remove or replace a pictogram that is already in the sequence below — every one of them must appear in your output, in the same order. Removing what is superfluous is the next step's job, not yours: a sequence that is still too long when you are done is expected.

HOW PICTOGRAM TRANSCRIPTION WORKS

A pictogram sequence is not a word-for-word rendering of the sentence. It carries the message with as few pictograms as the meaning allows: determiners, articles, partitives, auxiliaries, copulas and purely grammatical prepositions are never represented. Do not walk through the French sentence word by word looking for what is absent — add only a content concept the reader would need and that no pictogram in the sequence already carries.

WHAT EACH FIELD TELLS YOU

Each entry carries a `lemma`, and most carry a `definition` and a `description`.

- `lemma` is a short label. It is often ambiguous on its own and never enough to decide.
- `definition` states what the pictogram means. Use it to judge semantic fit: whether this sense of the lemma is the sense the sentence uses.
- `description` states what the pictogram shows. Use it to judge how a reader will actually read the pictogram, and to tell apart two entries whose definitions look equally plausible.

Read the definition first, then the description. A pictogram whose definition does not match the sentence is the wrong pictogram however well it is drawn; among pictograms whose definitions fit, the description decides.

Never use a definition or a description to invent concepts or attributes the text does not state.

ADD RULES

Read the sequence against the source text and check that nothing the sentence cannot be understood without is absent: an argument of the main predicate, the main verb, or a tense, negation or question marker the text requires. Add from the candidates what is missing. Never add a determiner, an article, a partitive, an auxiliary, a copula or a purely grammatical preposition — these are never part of a pictogram sequence. Never add a semantically empty verb to fill what looks like a missing predicate: in "avez-vous froid" what is predicated is the state itself, not the verb.
Most sequences need nothing added. Returning the sequence exactly as given is the expected outcome, not a failure. Only when a concept the sentence cannot be understood without is genuinely absent, and no candidate names it exactly, take a candidate rather than leaving a gap: a sequence missing its object says nothing at all, while an approximate pictogram in the right domain still tells the reader what the sentence is about. Choose in two steps: read the definitions to find which candidates come closest to the missing meaning, then among those take the one whose description best shows the situation the sentence describes. Use exactly one pictogram — never string several together to build up a concept, and never add an empty verb.

EXAMPLES

A missing concept restored:
Text: "elle a oublié son parapluie dans le train"
Sequence:
[{"lemma": "she", "picto_id": "260"}, {"lemma": "past", "picto_id": "261"}, {"lemma": "forget", "picto_id": "262"}, {"lemma": "in", "picto_id": "265"}, {"lemma": "train", "picto_id": "266"}]
Candidates:
[{"lemma": "she", "picto_id": "260"}, {"lemma": "past", "picto_id": "261"}, {"lemma": "forget", "picto_id": "262"}, {"lemma": "umbrella", "picto_id": "263"}, {"lemma": "in", "picto_id": "265"}, {"lemma": "train", "picto_id": "266"}]
Final:
[{"lemma": "she", "picto_id": "260"}, {"lemma": "past", "picto_id": "261"}, {"lemma": "forget", "picto_id": "262"}, {"lemma": "umbrella", "picto_id": "263"}, {"lemma": "in", "picto_id": "265"}, {"lemma": "train", "picto_id": "266"}]
(the object of the verb was missing; "umbrella" is added right after the verb, not appended at the end)

Output a single JSON list with the schema:
\ph|JSON schema: lemma + picto\_id^

Original text (in \ph|source language^):
\ph|source sentence^

Sequence:
\ph|current sequence^

Candidates:
\ph|retrieved candidates^
\end{Verbatim}
\end{promptbox}

\begin{promptbox}{RAPS Stage~3c --- Sequence composition: \textsc{Remove} (\texttt{gpt-oss-120b})}{prompt:raps-remove}
\begin{Verbatim}
You are correcting a pictogram sequence for Augmentative and Alternative Communication (AAC), one operation at a time. This is step 3 of 3: REMOVE, the final step. Output the complete, final sequence.

Remove any pictogram from the sequence that is not needed to convey the message. Never add, replace or reorder anything: every pictogram you keep must reappear in your output, unchanged, in its original relative order, using only picto_ids already present in the sequence below. You may not introduce any picto_id that is not already in the sequence.

REMOVE RULES

Remove a pictogram when:
a) it is an article, an auxiliary, a repeated marker, or a purely grammatical preposition carrying no meaning on its own ('de' in possession, 'à' before an indirect object).
    Never assume a preposition is redundant based only on its form; check whether removing it changes the meaning or grammatical role of its complement
    Be especially careful with "et", "ou", "mais", "par", "sans", "avec", "contre", "entre", "dans" and "à": never remove them unless you have verified that their removal does not change the meaning or grammatical structure.
   Possessives (mon, ton, son, notre, votre) are NOT in this category: they identify whose thing is meant and are always kept.
b) it is a verb carrying no meaning of its own, its sense being entirely in its complement. This covers copulas (le chat EST dans la boîte), auxiliaries, impersonal expressions (il FAIT chaud), and light-verb constructions (AVOIR faim, FAIRE attention, PRENDRE une décision). Three exceptions — keep the verb: when it states actual possession (as-tu un stylo), when it is causative (faire tomber, faire pleurer), and when removing it would leave the clause with no complement at all (tout VA bien). A complement that is present is never a reason to keep the verb: in "avez-vous froid" the state "cold" remains, so "have" goes; in "êtes-vous diabétique" the state "diabetes" remains, so "be" goes.
c) it has no basis in the text — it corresponds to no word, concept or entity actually mentioned. Never keep one because it merely fits the general topic.
d) its depiction would confuse or mislead the reader, and the concept is not essential.
e) the reader would still recover the message without it. Intensifiers, redundant qualifiers, and attributes already visible in another pictogram's depiction go here.
Never remove: the arguments of the main predicate (subject, object, key participants), the main verb unless it falls under b) above, negation, tense and question markers, or anything encoding location, direction or containment. The question marker is never redundant and should always be at the end, the best question lemma is "question": keep it even when the sentence already contains an interrogative word. Dropping a concept is a worse error than keeping one pictogram too many. When in doubt, keep it.

EXAMPLES

Shortening to the message:
Text: "le petit garçon mange rapidement une grosse pomme rouge"
Sequence:
[{"lemma": "small", "picto_id": "210"}, {"lemma": "boy", "picto_id": "211"}, {"lemma": "eat", "picto_id": "212"}, {"lemma": "quickly", "picto_id": "213"}, {"lemma": "big", "picto_id": "214"}, {"lemma": "red", "picto_id": "215"}, {"lemma": "apple", "picto_id": "216"}]
Kept:
[{"lemma": "boy", "picto_id": "211"}, {"lemma": "eat", "picto_id": "212"}, {"lemma": "apple", "picto_id": "216"}]
(the message is a boy eating an apple; "small", "quickly" and "big" are qualifiers the reader does not need, and "red" is already carried by the apple's own depiction)

A light verb that must be kept:
Text: "le vent a fait tomber les outils"
Sequence:
[{"lemma": "wind", "picto_id": "240"}, {"lemma": "past", "picto_id": "241"}, {"lemma": "make", "picto_id": "242"}, {"lemma": "fall", "picto_id": "243"}, {"lemma": "tool", "picto_id": "244"}]
Kept:
[{"lemma": "wind", "picto_id": "240"}, {"lemma": "past", "picto_id": "241"}, {"lemma": "make", "picto_id": "242"}, {"lemma": "fall", "picto_id": "243"}, {"lemma": "tool", "picto_id": "244"}]
(here "faire" is causative — the wind caused the falling, it did not fall itself — so "make" is kept; a causative verb carries the action and is never removed)

Collapsing a symptom split into two redundant pictograms:
Text: "as-tu des douleurs articulaires ?"
Sequence:
[{"lemma": "you", "picto_id": "220"}, {"lemma": "have", "picto_id": "221"}, {"lemma": "joint pain", "picto_id": "222"}, {"lemma": "joint", "picto_id": "223"}, {"lemma": "question", "picto_id": "224"}]
Kept:
[{"lemma": "you", "picto_id": "220"}, {"lemma": "joint pain", "picto_id": "222"}, {"lemma": "question", "picto_id": "224"}]
("joint pain" already depicts the joint; "joint" on its own adds nothing the reader does not already get from "joint pain", so it falls under e); "have" is a light verb whose state — the pain — remains, so it falls under b) exactly like "avez-vous froid")

A light verb kept by mistake:
Text: "as-tu des sueurs nocturnes ?"
Sequence:
[{"lemma": "you", "picto_id": "230"}, {"lemma": "have", "picto_id": "231"}, {"lemma": "night sweat", "picto_id": "232"}, {"lemma": "question", "picto_id": "233"}]
Kept:
[{"lemma": "you", "picto_id": "230"}, {"lemma": "night sweat", "picto_id": "232"}, {"lemma": "question", "picto_id": "233"}]
("avoir des sueurs nocturnes" is a light-verb construction like "avoir froid": the state "night sweat" remains on its own, so "have" goes)

Output a single JSON list containing only the pictograms to KEEP, in their original order, using the schema:
\ph|JSON schema: lemma + picto\_id^

Original text (in \ph|source language^):
\ph|source sentence^

Sequence:
\ph|current sequence^
\end{Verbatim}
\end{promptbox}

\subsubsection{LLM-as-a-Judge evaluation of instruct answers}
\label{appendix:prompts-judge}

Answers of the instruct model are scored by an LLM judge
(\texttt{judge-model}) 
on three criteria, each in a separate call:
\emph{accuracy} (Prompt~\ref{prompt:judge-exactitude}),
\emph{relevance} (Prompt~\ref{prompt:judge-pertinence}) and
\emph{simplicity} (Prompt~\ref{prompt:judge-simplicite}).
All three share the same template and differ only in their scoring rubric.
The judge receives the question, the answer to evaluate and a reference
answer scored 5, writes a short comment, then outputs an integer score
from 1 to 5. Prompts are given in French, as used at run time.

\begin{promptbox}{LLM-as-a-Judge --- Accuracy (\textit{Exactitude})}{prompt:judge-exactitude}
\begin{Verbatim}
### Description de la tâche
On te donne une question, une réponse à évaluer, une réponse de référence qui obtient la note de 5, et une grille de notation représentant un critère d'évaluation.
La réponse a été produite par un assistant destiné à des personnes qui utilisent la communication alternative et augmentée (CAA) : personnes avec un trouble du langage, une déficience intellectuelle, un trouble du spectre de l'autisme, une aphasie, etc. Elles posent des questions courtes ; l'assistant doit répondre en une ou deux phrases très simples, qu'elles peuvent comprendre seules.
1. Rédige un commentaire (2 à 4 phrases) qui évalue la qualité de la réponse strictement selon la grille de notation donnée, et non de manière générale, en citant des éléments précis de la réponse.
2. Donne ensuite une note entière entre 1 et 5, en te référant à la grille.
3. Renvoie uniquement un objet JSON de la forme :
   {"commentaire": "<ton commentaire>", "note": <entier entre 1 et 5>}
4. Ne génère aucun texte en dehors de cet objet JSON.

### Question
\ph|question^

### Réponse à évaluer
\ph|answer to evaluate^

### Réponse de référence (note 5)
\ph|reference answer^

### Grille de notation
Exactitude : [La réponse est-elle factuellement correcte, sans information fausse ni inventée ? La réponse de référence est un exemple de bonne réponse, pas la seule réponse valide : ne pénalise pas une réponse correcte formulée autrement ou plus courte, ni une information juste absente de la référence. Une simplification normale du langage FALC n'est pas une erreur. Si la réponse n'affirme aucun fait (par exemple, elle renvoie seulement vers un proche), juge uniquement ce qu'elle affirme. Ne juge ici ni la pertinence, ni la simplicité de la langue.]
Note 1 : La réponse contredit la référence sur l'essentiel, ou contient une information fausse, inventée ou dangereuse (fait, chiffre, nom, lieu, date, procédure, conseil de santé).
Note 2 : La réponse contient une erreur nette sur un point secondaire, ou mêle des informations justes et des informations fausses.
Note 3 : La réponse ne contient pas d'erreur franche, mais une imprécision ou une approximation qui peut tromper l'utilisateur.
Note 4 : Tout est correct, avec une légère approximation sans conséquence pour l'utilisateur.
Note 5 : Tout ce que la réponse affirme est correct et cohérent avec la référence ; rien n'est inventé.

### Évaluation (JSON) :
\end{Verbatim}
\end{promptbox}

\begin{promptbox}{LLM-as-a-Judge --- Relevance (\textit{Pertinence})}{prompt:judge-pertinence}
\begin{Verbatim}
### Description de la tâche
On te donne une question, une réponse à évaluer, une réponse de référence qui obtient la note de 5, et une grille de notation représentant un critère d'évaluation.
La réponse a été produite par un assistant destiné à des personnes qui utilisent la communication alternative et augmentée (CAA) : personnes avec un trouble du langage, une déficience intellectuelle, un trouble du spectre de l'autisme, une aphasie, etc. Elles posent des questions courtes ; l'assistant doit répondre en une ou deux phrases très simples, qu'elles peuvent comprendre seules.
1. Rédige un commentaire (2 à 4 phrases) qui évalue la qualité de la réponse strictement selon la grille de notation donnée, et non de manière générale, en citant des éléments précis de la réponse.
2. Donne ensuite une note entière entre 1 et 5, en te référant à la grille.
3. Renvoie uniquement un objet JSON de la forme :
   {"commentaire": "<ton commentaire>", "note": <entier entre 1 et 5>}
4. Ne génère aucun texte en dehors de cet objet JSON.

### Question
\ph|question^

### Réponse à évaluer
\ph|answer to evaluate^

### Réponse de référence (note 5)
\ph|reference answer^

### Grille de notation
Pertinence : [La réponse répond-elle directement à la question posée, et adopte-t-elle le bon comportement entre répondre directement et renvoyer vers une personne ? Quand la réponse dépend de la situation de la personne (santé, école, famille…), un renvoi clair vers la bonne personne (médecin, enseignant, proche…) compte comme une réponse directe. Ne juge ici ni l'exactitude détaillée des faits, ni la simplicité de la langue.]
Note 1 : La réponse est hors sujet, répond à une autre question ou esquive la question sans raison.
Note 2 : La réponse effleure la question sans vraiment y répondre ou reste vague, ou renvoie vers une personne alors que la question porte sur une connaissance générale, ou donne un conseil personnalisé alors que la réponse dépend de la personne, du lieu ou du moment.
Note 3 : La réponse traite la question, mais de façon indirecte ou incomplète, ou le renvoi vers une personne est vague (« demande à quelqu'un »).
Note 4 : La réponse traite directement la question, avec un défaut mineur : un détail hors sujet qui peut distraire, ou une première phrase qui tarde à répondre.
Note 5 : La réponse traite la question dès la première phrase et tout ce qu'elle dit sert à répondre ; elle répond directement pour une connaissance générale, ou renvoie clairement vers la bonne personne quand la réponse en dépend.

### Évaluation (JSON) :
\end{Verbatim}
\end{promptbox}

\begin{promptbox}{LLM-as-a-Judge --- Simplicity (\textit{Simplicité})}{prompt:judge-simplicite}
\begin{Verbatim}
### Description de la tâche
On te donne une question, une réponse à évaluer, une réponse de référence qui obtient la note de 5, et une grille de notation représentant un critère d'évaluation.
La réponse a été produite par un assistant destiné à des personnes qui utilisent la communication alternative et augmentée (CAA) : personnes avec un trouble du langage, une déficience intellectuelle, un trouble du spectre de l'autisme, une aphasie, etc. Elles posent des questions courtes ; l'assistant doit répondre en une ou deux phrases très simples, qu'elles peuvent comprendre seules.
1. Rédige un commentaire (2 à 4 phrases) qui évalue la qualité de la réponse strictement selon la grille de notation donnée, et non de manière générale, en citant des éléments précis de la réponse.
2. Donne ensuite une note entière entre 1 et 5, en te référant à la grille.
3. Renvoie uniquement un objet JSON de la forme :
   {"commentaire": "<ton commentaire>", "note": <entier entre 1 et 5>}
4. Ne génère aucun texte en dehors de cet objet JSON.

### Question
\ph|question^

### Réponse à évaluer
\ph|answer to evaluate^

### Réponse de référence (note 5)
\ph|reference answer^

### Grille de notation
Simplicité : [La réponse est-elle facile à lire et à comprendre (FALC) pour une personne qui utilise la CAA ? Juge uniquement la forme du texte, pas son contenu : ne te demande pas si la réponse est juste ou utile. La réponse de référence montre la longueur et le niveau de langue attendus.]
Note 1 : Le texte est très difficile à comprendre : texte long, phrases longues et complexes, vocabulaire technique ou soutenu, plusieurs idées mêlées.
Note 2 : Le texte est difficile à comprendre : plusieurs défauts parmi jargon ou sigle, phrases de plus de 20 mots, subordonnées enchaînées, plus de trois phrases.
Note 3 : Le texte se comprend avec effort : un défaut net (un mot technique ou abstrait, une phrase longue, une voix passive, une négation double, ou trois phrases).
Note 4 : Le texte est simple, avec un défaut mineur : un mot un peu moins courant, une phrase un peu longue, ou deux idées dans la même phrase.
Note 5 : Le texte est conforme au FALC : une ou deux phrases courtes (environ 12 mots au plus chacune), une seule idée par phrase, des mots courants de la vie quotidienne, sans sigle, jargon, mot abstrait ni tournure complexe.

### Évaluation (JSON) :
\end{Verbatim}
\end{promptbox}
\subsection{Details Results LLM as Judge}
\label{appendix:llm_judge_details}
\begin{table*}[!h]
  \centering
  \footnotesize
  \begin{tabular}{l cccc}
    \hline
    \textbf{Judge} & \textbf{Rel.} & \textbf{Corr.} & \textbf{Simp.} & \textbf{Overall} \\
    \hline
    \multicolumn{5}{l}{\textbf{Ministral 0-shot}} \\
    Mistral-Small & 0.793 & 0.904 & 0.766 & 0.821 \\
    GPT-OSS       & 0.970 & 0.912 & 0.686 & 0.856 \\
    DeepSeek-R1   & 0.818 & 0.924 & 0.743 & 0.828 \\
    Mean          & \textit{0.861} & \textit{0.913} & \textit{0.732} & \textit{0.835} \\
    \hline
    \multicolumn{5}{l}{\textbf{Ministral 8-shot}} \\
    Mistral-Small & 0.782 & 0.898 & 0.797 & 0.825 \\
    GPT-OSS       & 0.958 & 0.901 & 0.842 & 0.900 \\
    DeepSeek-R1   & 0.811 & 0.902 & 0.783 & 0.832 \\
    Mean          & \textit{0.850} & \textit{0.900} & \textit{0.807} & \textit{0.852} \\
    \hline
    \multicolumn{5}{l}{\textbf{\panta-v2}} \\
    Mistral-Small & 0.770 & 0.893 & 0.807 & 0.823 \\
    GPT-OSS       & 0.960 & 0.884 & 0.858 & 0.901 \\
    DeepSeek-R1   & 0.809 & 0.892 & 0.803 & 0.835 \\
    Mean          & \textit{0.846} & \textit{0.890} & \textit{0.823} & \textit{0.853} \\
    \hline
    \multicolumn{5}{l}{\textbf{\panta-v2 (picto)}} \\
    Mistral-Small & 0.624 & 0.722 & 0.718 & 0.688 \\
    GPT-OSS       & 0.752 & 0.747 & 0.859 & 0.786 \\
    DeepSeek-R1   & 0.678 & 0.761 & 0.789 & 0.743 \\
    Mean          & \textit{0.685} & \textit{0.743} & \textit{0.789} & \textit{0.739} \\
    \hline
  \end{tabular}
  \caption{Per-judge LLM-judge scores on the 1,285 test questions. Scores from 1 to 5 are rescaled to $[0,1]$. Mean: average over the judges that returned a score.}
  \label{tab:judge-detail}
\end{table*}

\twocolumn[{%
  \subsection{Pictogram evaluation interface}
  \label{appendix:interface}
  \centering
  \includegraphics[width=\textwidth,height=0.9\textheight,keepaspectratio]{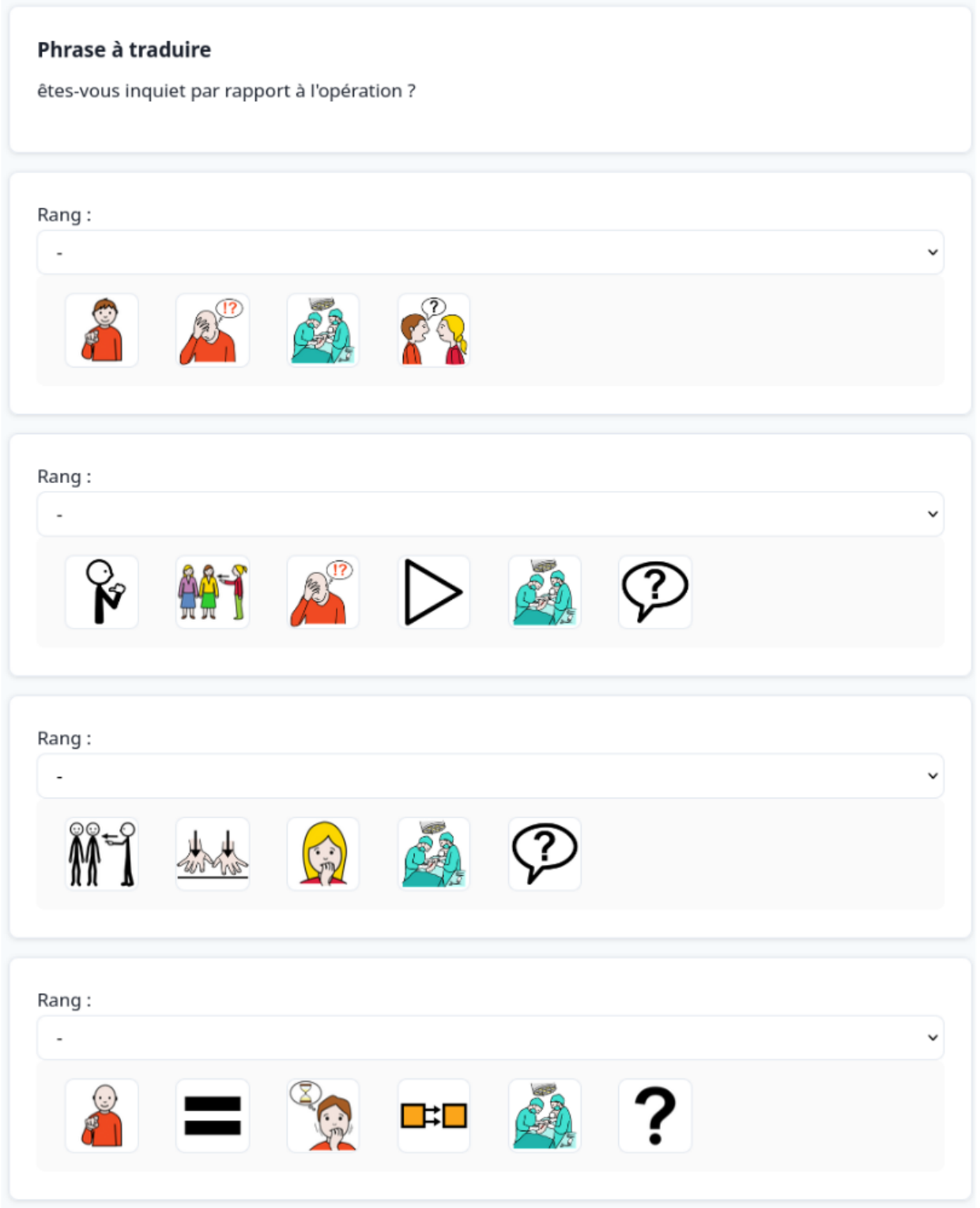}
  \captionof{figure}{Pictogram evaluation interface}
  \label{app:interface}
  \vspace{1em}
}]
\subsection{Human Evaluation Protocol for the Instruction-Following Model}
\label{app:human-instruct}

\begin{figure}[t]
\centering
\includegraphics[width=\columnwidth]{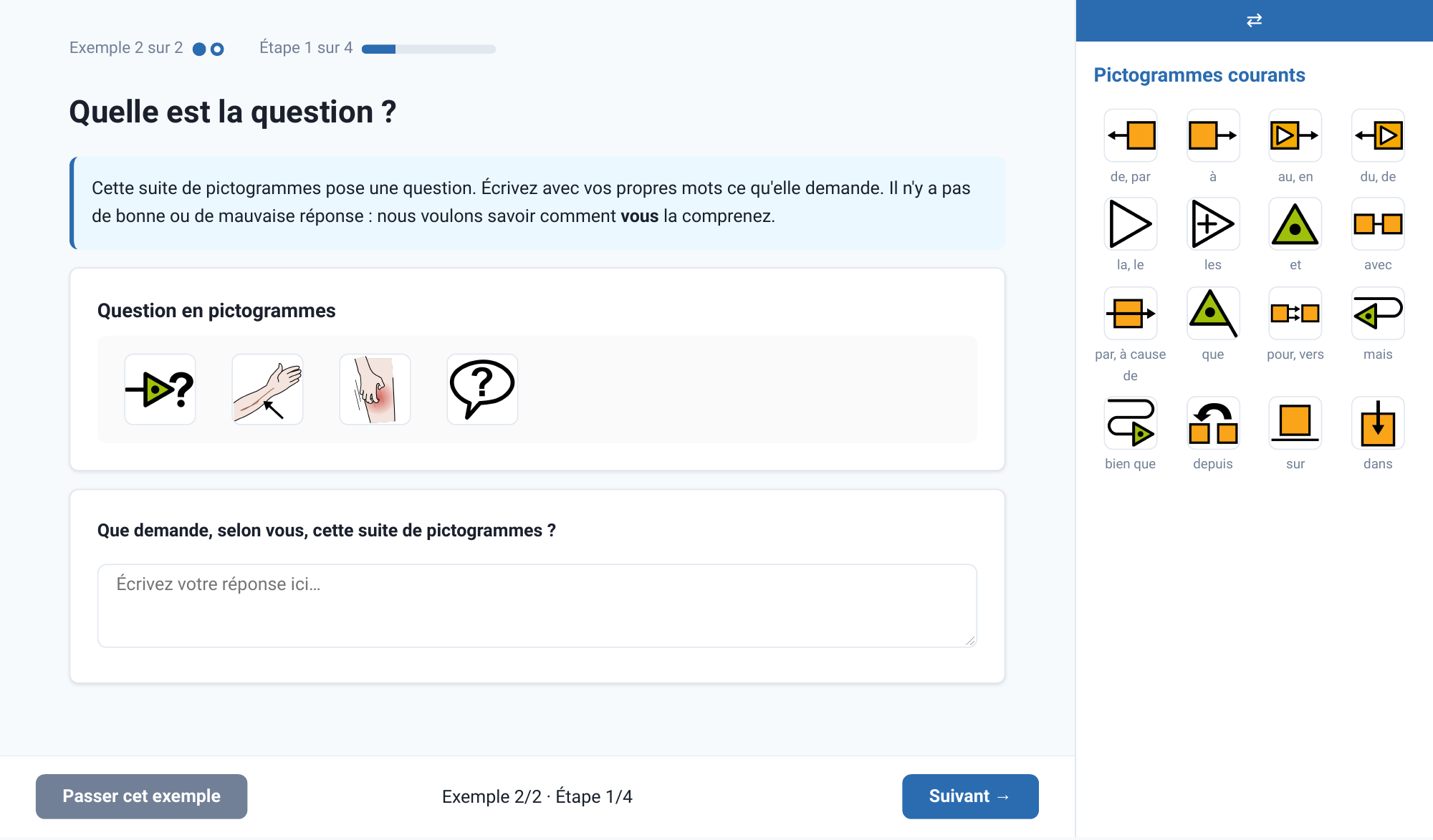}
\caption{Annotation interface, step 1: transcription of the pictogram sequence
into French, without any textual context.}
\label{fig:annot-transcription}
\end{figure}

\begin{figure}[t]
\centering
\includegraphics[width=\columnwidth]{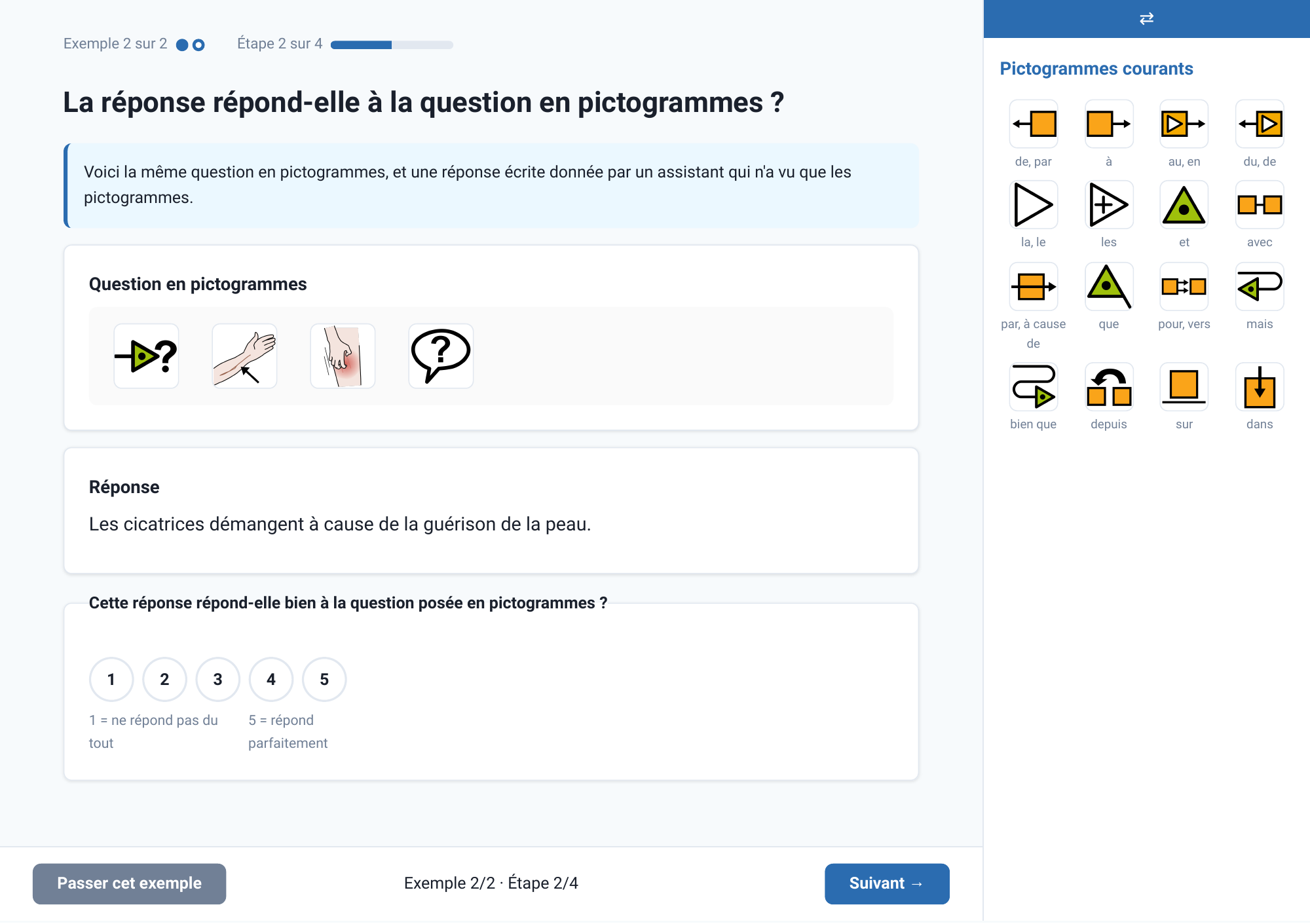}
\caption{Annotation interface, step 2: adequacy of the answer to the question.}
\label{fig:annot-adequacy}
\end{figure}

\begin{figure}[t]
\centering
\includegraphics[width=\columnwidth]{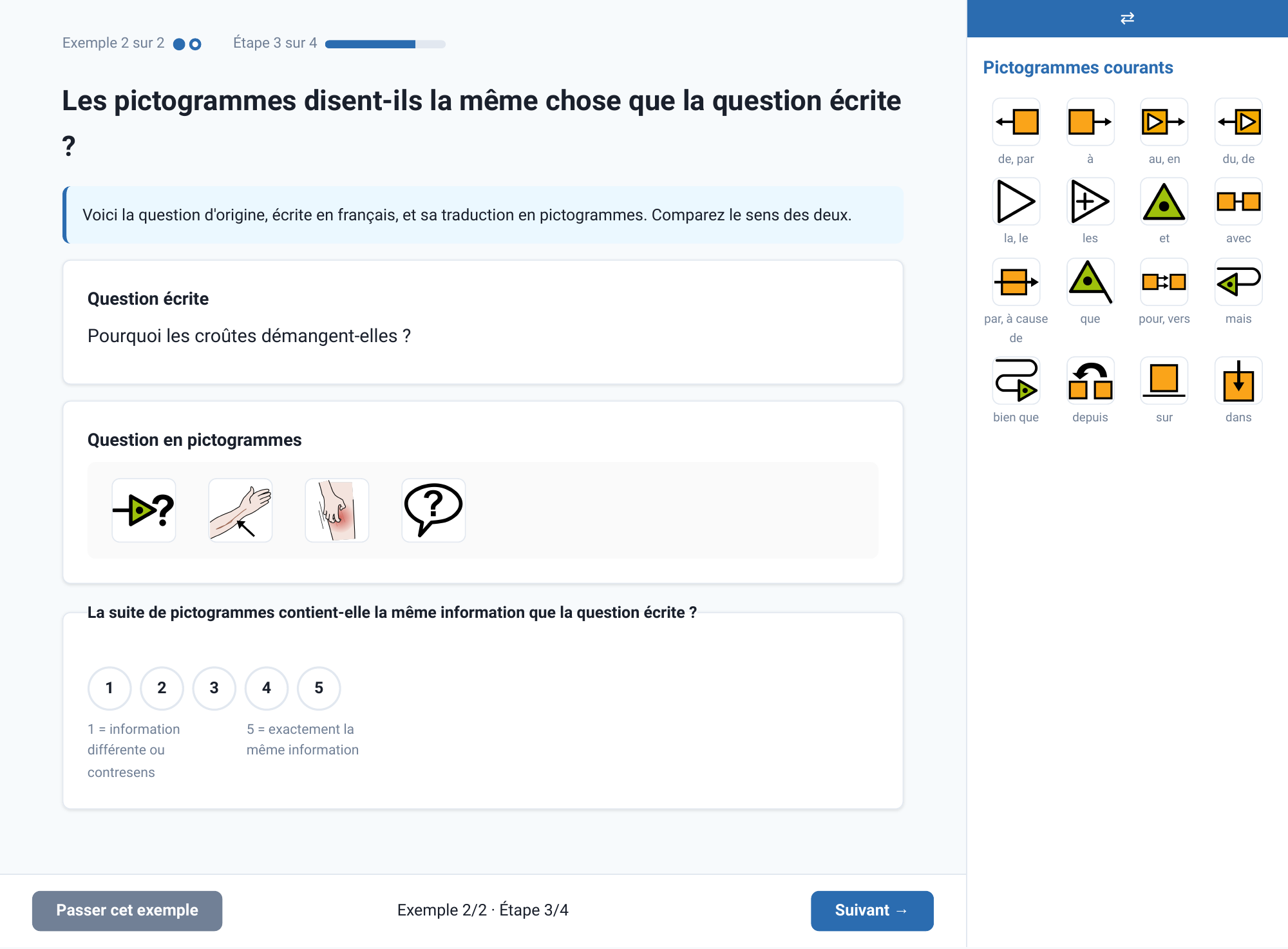}
\caption{Annotation interface, step 3: consistency of the pictogram answer
with the text answer, or fidelity of the pictogram question to the original
question.}
\label{fig:annot-reference}
\end{figure}

\begin{figure}[t]
\centering
\includegraphics[width=\columnwidth]{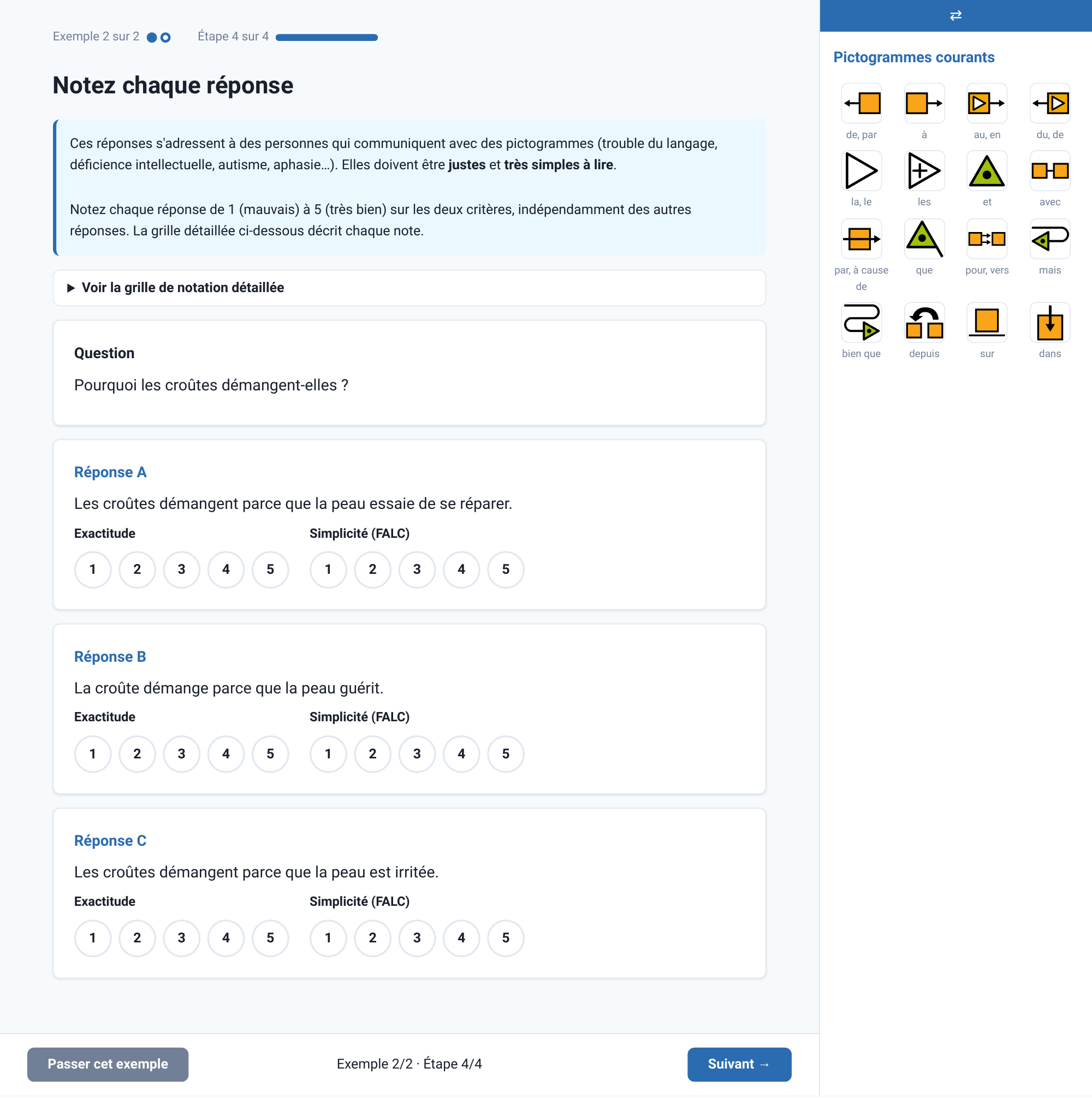}
\caption{Annotation interface, step 4: correctness and simplicity of text
answers to text questions.}
\label{fig:annot-text}
\end{figure}

The human evaluation focuses on cross-modal settings: text-to-text answers are
already evaluated with reference-based metrics and LLM judges
(Section~\ref{sec:auto-eval}), whereas pictogram-to-pictogram interactions are
difficult to evaluate reliably. Annotators use a web interface in four steps
(Figures~\ref{fig:annot-transcription}--\ref{fig:annot-text}). All ratings are
given on a 1--5 scale and mapped to $[0,1]$ as $(s-1)/4$.

\paragraph{Step 1: transcription.} Annotators first see only a pictogram
sequence, either the pictogram answer of \textit{\panta-v2} (text question,
pictogram answer) or the pictogram question produced by RAPS (pictogram
question, text answer), and transcribe it into French
(Figure~\ref{fig:annot-transcription}). This forces them to interpret the
sequence on its own, without the context of the textual question or answer.

\paragraph{Step 2: adequacy.} Annotators then see the textual question (text
question, pictogram answer) or the text answer of \textit{\panta-v2}
(pictogram question, text answer), and rate whether the answer addresses the
question (Figure~\ref{fig:annot-adequacy}). This step measures whether the
model answers correctly in the target modality.

\paragraph{Step 3: transcription quality.} Finally, annotators see the text
that the pictogram sequence is meant to convey: the text answer generated by
\textit{\panta-v2} for the same question, or the original textual question
transcribed by RAPS (Figure~\ref{fig:annot-reference}). They rate whether the
pictogram sequence preserves its meaning (\textit{transcription quality}).
Comparing this score with correctness shows whether a low correctness comes
from the model or from the reading of the pictogram sequence.

\paragraph{Step 4: text answers.} To validate the LLM judges, annotators rate
text answers to text questions for correctness and simplicity
(Figure~\ref{fig:annot-text}). Answers from \textit{\panta-v2}, Ministral-3-8B
(0-shot) and the reference answers are shown without system names.

\providecommand{\licTODO}{\textcolor{red}{\textbf{[TO FILL]}}}

\subsection{Licenses of Scientific Artifacts}
\label{sec:licenses}

Tables~\ref{tab:licenses-models} and~\ref{tab:licenses-data} list the licenses of all models, tools and datasets used in this work. All artifacts were used for non-commercial research purposes, consistent with their licenses and intended uses. Since our pictogram sequences are built on the ARASAAC vocabulary, which is distributed under CC BY-NC-SA 4.0, the text-pictogram resources we release inherit the same non-commercial, share-alike terms. The \anonymousdataset source is not redistributed until approved by our legal department; this is why only 300k of the 1.4M pairs are released.

\begin{table}[h]
\centering
\small
\begin{tabular}{@{}ll@{}}
\toprule
\textbf{Model / Tool} & \textbf{License} \\
\midrule
\multicolumn{2}{@{}l}{\textit{Backbone and RAPS pipeline}} \\
Ministral-3-8B-Instruct-2512 (inc.\ Pixtral) & Apache 2.0 \\
gpt-oss-120b & Apache 2.0 \\ 
Qwen3-32B & Apache 2.0 \\ 
Qwen3-14B & Apache 2.0 \\ 
Qwen3-Embedding-0.6B & Apache 2.0 \\ 
nllb-200-distilled-600M & CC BY-NC 4.0 \\ 
Stanza & Apache 2.0 \\ 
\midrule
\multicolumn{2}{@{}l}{\textit{LLM judges}} \\
Mistral-Small-3.2-24B-Instruct-2506 & Apache 2.0 \\ 
DeepSeek-R1-Distill-Qwen-32B & MIT \\ 
\midrule
\multicolumn{2}{@{}l}{\textit{Speech}} \\
Voxtral-4B-TTS-2603 & CC BY-NC 4.0 \\
BiCodec / Spark-TTS-0.5B (weights) & CC BY-NC-SA 4.0 \\ 
Silero VAD & MIT \\ 
pyannote.audio (diarization) & MIT \\ 
whisper-large-v3 / large-v3-turbo & MIT \\ 
Canary-1b-v2 & CC BY 4.0 \\ 
ssl-music-detection-music2vec & Research agreement\\
UTMOS-v2 & MIT License \\
spkrec-ecapa-voxceleb & Apache 2.0 \\ 
\midrule
\multicolumn{2}{@{}l}{\textit{Evaluation}} \\
CamemBERT (BERTScore) & MIT \\ 
\bottomrule
\end{tabular}
\caption{Licenses of the models and tools used in this work.}
\label{tab:licenses-models}
\end{table}

\begin{table}[h]
\centering
\small
\begin{tabular}{@{}ll@{}}
\toprule
\textbf{Dataset / Resource} & \textbf{License} \\
\midrule
\multicolumn{2}{@{}l}{\textit{Pictogram vocabulary}} \\
ARASAAC & CC BY-NC-SA 4.0 \\
WordNet & WordNet License \\ 
\midrule
\multicolumn{2}{@{}l}{\textit{Text sources (pre-training corpus)}} \\
Vikidia & CC-BY-SA 3.0 \\ 
Cosmopedia & Apache 2.0 \\ 
Project Gutenberg &  Project Gutenberg License \\ 
\anonymousdataset & Not redistributed \\
Propicto & cc0-1.0 \\
\midrule
\multicolumn{2}{@{}l}{\textit{Instruction data}} \\
French-Instruct & MIT \\
\midrule
\multicolumn{2}{@{}l}{\textit{Speech data}} \\
MLS (French) & CC BY 4.0 \\ 
EPAC & Commercial Use - ELRA VAR \\
Audiocite.net & CC BY-NC  \\
YODAS & CC BY 3.0 \\ 
\midrule
\multicolumn{2}{@{}l}{\textit{Human-annotated evaluation corpora}} \\
Medical (BabelDr) & CC BY-NC-SA 4.0 \\
Email (WAI-NOT) & CC BY-NC-SA 4.0 \\
Book & CC BY-NC-SA 4.0 \\
\end{tabular}
\caption{Licenses of the datasets and resources used and released in this work.}
\label{tab:licenses-data}
\end{table}

\end{document}